\documentclass[a4paper,fleqn]{cas-dc}

\usepackage[authoryear,longnamesfirst]{natbib}

\usepackage{amsmath,amssymb}
\usepackage{booktabs}
\usepackage{multirow}
\usepackage{tikz}
\usetikzlibrary{arrows.meta,positioning}
\usepackage{array} 

\newcommand{\cmark}{\checkmark}
\newcommand{\xmark}{$\times$}
\newcommand{\trimark}{$\triangle$}

\usepackage{cas-common}

\shorttitle{Open-Loop WWTP Simulation for Digital-Twin Operator Support}
\shortauthors{Simethy et~al.}

\hypersetup{
  pdftitle={Data-Driven Open-Loop Simulation for Digital-Twin Operator Decision Support in Wastewater Treatment},
  pdfauthor={Gary Simethy, Daniel Ortiz Arroyo, Petar Durdevic}
}

\ExplSyntaxOn
\bool_gset_true:N \g_stm_nologo_bool
\RenewDocumentCommand \printorcid {} {}
\cs_gset:Npn \__first_footerline:
  {
    \group_begin:
    \small
    \sffamily
    \ifnum\theblind>0\relax
    \else
      \__short_authors: :~
    \fi
    { \rmfamily \itshape arXiv~preprint }
    \group_end:
  }
\ExplSyntaxOff

\begin{document}

\title[mode=title]{Data-Driven Open-Loop Simulation for Digital-Twin Operator Decision Support in Wastewater Treatment}

\author[1]{Gary Simethy}
\cormark[1]
\ead{gasi@energy.aau.dk}

\author[1]{Daniel Ortiz Arroyo}

\author[1]{Petar Durdevic}

\affiliation[1]{organization={Aalborg University, Department of Energy},
  addressline={Niels Bohrs Vej 8},
  city={Esbjerg},
  postcode={DK-6700},
  country={Denmark}}

\cortext[cor1]{Corresponding author.}

\begin{abstract}
Wastewater treatment plants (WWTPs) need digital-twin-style decision support tools
that can simulate plant response under prescribed control plans, tolerate irregular
and missing sensing, and remain informative over 12--36 h planning horizons. Meeting
these requirements with full-scale plant data remains an open engineering-AI
challenge. We present CCSS-RS, a controlled continuous-time state-space model that
separates historical state inference from future control and exogenous rollout. The
model combines typed context encoding, gain-weighted forcing of prescribed and
forecast drivers, semigroup-consistent rollouts, and Student-$t$ plus hurdle outputs
for heavy-tailed and zero-inflated WWTP sensor data. On the public Aved{\o}re
full-scale benchmark, with 906,815 timesteps, 43\% missingness, and 1--20 min
irregular sampling, CCSS-RS achieves RMSE 0.696 and CRPS 0.349 at $H=1000$ across
10,000 test windows. This reduces RMSE by 40--46\% relative to Neural CDE baselines
and by 31--35\% relative to simplified internal variants. Four case studies using a
frozen checkpoint on test data demonstrate operational value: oxygen-setpoint
perturbations shift predicted ammonium by $-2.3$ to $+1.4$ over horizons 300--1000;
a smoothed setpoint plan ranks first in multi-criterion screening; context-only
sensor outages raise monitored-variable RMSE by at most 10\%; and ammonium, nitrate,
and oxygen remain more accurate than persistence throughout the rollout. These
results establish CCSS-RS as a practical learned simulator for offline scenario
screening in industrial wastewater treatment, complementary to mechanistic models.
\end{abstract}


\maketitle

\section{Introduction}
\label{sec:intro}

Effective process control in wastewater treatment depends on anticipating how plant behaviour will respond to changing operating conditions and planned control actions.
Consider the task facing a wastewater treatment plant operator before the next shift:
compare two future aeration plans, estimate how $\text{NH}_4$, $\text{NO}_3$,
$\text{N}_2\text{O}$, and dissolved oxygen will evolve over the next 12--36 hours, and
identify whether the plan differences are large enough to matter operationally. This is
not classical forecasting. In open-loop simulation, future control inputs are known by
construction and must drive system evolution; the unknown quantity is the resulting future
process state. This capability is central to digital-twin-style what-if analysis,
scenario screening, and model predictive control (MPC) in industrial automation
\citep{camacho2007model,qin2003survey}. Data-driven simulators that embed this
capability directly from plant historian data represent a practical route to operator
decision support without requiring full mechanistic model recalibration, and are
increasingly relevant to smart-water and water Industry~4.0 initiatives
\citep{newhart2019data,garridobaserba2020fourth}.

Wastewater treatment plants make this problem especially demanding. Full-scale facilities
generate irregular and asynchronous sensor streams with substantial missingness
\citep{hansen2024avedore}, exhibit complex multi-phase biological dynamics under alternating
operating regimes \citep{hauduc2015activated,dotro2011intermittent}, and require long
prediction horizons for planning and MPC-style screening, which under fine-grained sampling
can correspond to hundreds to thousands of rollout steps. Classical Activated Sludge Model
(ASM) family models \citep{henze2000asm} remain indispensable for mechanism-level analysis,
but they require extensive calibration and may be sensitive to plant-specific conditions and
data quality issues \citep{gernaey2004activated}. Purely data-driven one-step predictors
often degrade when repeatedly rolled out over long horizons because forecast errors compound
across steps \citep{ben2012review}. In open-loop industrial scenario analysis, the simulator
must also remain explicitly conditioned on the prescribed future control trajectory. The
engineering need is therefore not another benchmark forecaster, but a learned simulator that
can use realistic plant data for operational support without pretending to replace
mechanistic process understanding.

Recent AI models for long-horizon time series improve standard forecasting benchmarks, but
most are not designed for controlled industrial scenario screening. They often assume
regular, synchronised sensor grids or require asynchronous plant signals to be resampled
onto such grids. They also tend to treat future covariates as generic predictive context or
ignore the distinction between control decisions and process states. This leaves a gap
between modern sequence modelling and the requirements of application-facing industrial
decision-support systems. This paper addresses that gap with an architecture whose core
design choices, including typed historical context encoding, explicit separation of future
controls from inferred states, continuous-time $\Delta t$-aware scans, and
control-conditioned rollout forcing, are matched to real-plant data conditions and
evaluated through application-oriented case studies that reflect actual operator workflows.

We present CCSS-RS (Continuous Controlled State-Space Model with Regime Switching), a
learned simulator for open-loop wastewater process analysis. The model explicitly
separates state variables $x(t)$ from control inputs $u(t)$ and exogenous variables
$w(t)$ under continuous-time dynamics $\dot{x}(t)=f(x(t),u(t),w(t),t)$.
Typed context encoding allows irregular observations to be consumed without resampling or
imputation, a gain-weighted forcing mechanism injects prescribed controls and forecast exogenous
drivers into latent dynamics, regime-specific experts capture phase-dependent behaviour
\citep{teh2006hierarchical,fox2011sticky}, and semigroup consistency improves
step-size-robust rollout under irregular sampling \citep{chen2023deeposg}. Each design
choice serves an application goal: a data-driven simulator that can answer
digital-twin-style what-if questions under the data conditions of real full-scale WWTP
operation, requiring no process model recalibration and no signal resampling.

We evaluate CCSS-RS (6.9M parameters) against Neural CDE-Small (0.67M) and Neural
CDE-Large (4.1M) as the most directly compatible established baselines under the no-resampling WWTP
protocol, and against two simplified variants, CCSS-SR (7.3M, no regime switching) and
CCSS-G (7.3M, Gaussian likelihood) as controlled ablative comparisons. Despite having
more parameters, both simplified variants perform substantially worse, indicating that
the gap cannot be explained by model capacity alone. All models are evaluated on the
public Aved{\o}re WWTP dataset \citep{hansen2024avedore}, comprising 906,815 timesteps over two
years, 5 state variables, 6 control variables, 43\% overall missingness, and $CV=0.32$
in sampling intervals, across 10,000 evaluation windows.

Our contributions are fourfold: (i)~we frame long-horizon open-loop wastewater simulation
as an engineering-AI problem for industrial digital-twin decision support under known
future controls, irregular sampling, and substantial missingness, clarifying why generic
forecasting architectures are insufficient for this setting; (ii)~we present CCSS-RS, a
control-conditioned continuous-time state-space architecture with explicit state/control
separation, gain-weighted forcing, regime-aware dynamics, and semigroup-consistent
rollout, together with an analysis of what each component contributes at long horizons;
(iii)~we benchmark the approach on the public Aved{\o}re full-scale dataset at $H=1000$
under 43\% missingness, achieving RMSE 0.696 and CRPS 0.349 and outperforming the most directly compatible external baselines as well as simplified internal variants; and
(iv)~we provide four application-oriented case studies on test data only, showing how the
frozen simulator can be used for what-if plan comparison, multi-criterion candidate-plan
screening, context-outage robustness assessment, and empirical decision-horizon
interpretation, translating benchmark performance into concrete operator support value.

The novelty of CCSS-RS is both architectural and application driven. Methodologically, the
model occupies a design space between Neural CDEs, generic state-space forecasting
backbones, and RL-style world models, but it differs from each in how it represents
industrial control. Neural CDEs handle irregular timestamps, but usually combine
observations, controls, and exogenous variables into a common driving path
\citep{kidger2020neural}. Generic state-space models such as S4 and Mamba provide
efficient long-sequence modelling, but typically treat future controls as ordinary input
channels rather than as prescribed intervention trajectories
\citep{gu2022s4,gu2024mamba}. RL world models separate actions and observations, but are
primarily designed for closed-loop episodic learning rather than open-loop scenario
screening from irregular plant historian data
\citep{hafner2020dreamer,hafner2023dreamerv3}. CCSS-RS introduces an architecture tailored
to this industrial setting: typed context encoding separates inferred process states,
controllable inputs, and exogenous drivers; gain-weighted forcing injects prescribed and
forecast future drivers directly into the rollout; regime-specific experts model
phase-dependent WWTP dynamics; and semigroup consistency improves robustness to irregular
step sizes over long horizons. This makes the model a data-driven simulator for evaluating
prescribed operating plans, rather than a generic forecaster that extrapolates future
measurements.
\section{Related Work}
\label{sec:related}

\subsection{Forecasting Architectures and Their Control Limitations}

Recent long-horizon forecasting models demonstrate strong benchmark performance, but they
largely target problems with partly unknown future inputs and regular sampling. Transformer families such as Informer,
Autoformer, PatchTST, iTransformer, TFT, and TimeXer
\citep{zhou2021informer,wu2021autoformer,nie2023patchtst,liu2024itransformer,lim2021temporal,wang2024timexer}
improve generic long-range prediction, yet even the variants that accept future covariates generally do not make the engineering
distinction between a control plan that is known by construction and a process state that must
be inferred explicitly in the model structure. This limits direct transfer to
open-loop wastewater scenario screening, where future controls must drive the rollout.

State-space models are a more natural starting point because $\dot{x}=Ax+Bu$ mirrors
classical control structure \citep{kalman1960new,ljung1999system}. However, modern SSMs
such as S4, Mamba, and MambaTS \citep{gu2022s4,gu2024mamba,cai2024mambats} are generally evaluated on regularly sampled datasets and are not typically designed around
future control plans as privileged rollout drivers. Neural CDEs \citep{kidger2020neural} therefore
remain the closest external baseline for our setting because they natively handle
irregular sampling and continuous-time control paths; recent variants such as Log-NCDE
and Stable Neural SDE \citep{walker2024logncde,oh2024stable} strengthen this family
further. Their limitation for open-loop wastewater simulation is not compatibility but
representation: past observations, future controls, and exogenous forecasts are folded
into a single driving signal rather than being separated into inferred state and
prescribed plan.

\subsection{Missing Data, Irregular Sampling, and World Models}

Methods for incomplete process data fall into three broad groups. Explicit-missingness
models such as GRU-D \citep{che2018grud} expose masks and time gaps directly to the
sequence model; learned-imputation methods such as BRITS \citep{cao2018brits} and SAITS
\citep{du2023saits} reconstruct missing values before or during prediction; and
continuous-time models such as T-LSTM \citep{baytas2017tlstm}, Latent ODE
\citep{rubanova2019latent}, and GRU-ODE-Bayes \citep{brouwer2019gruode} evolve latent
states between irregular observations. These methods address missingness and irregularity,
but their primary objective is reconstruction, interpolation, or prediction rather than
forward simulation under a specified future plan. CCSS-RS builds on the continuous-time
lineage while adding explicit control handling, probabilistic belief tracking, and
regime-aware dynamics.

The world model paradigm from reinforcement learning \citep{ha2018world} offers the
conceptually closest parallel to our goal: PlaNet \citep{hafner2019planet} and Dreamer
\citep{hafner2020dreamer,hafner2021dreamerv2,hafner2023dreamerv3} use recurrent SSMs to
learn latent dynamics conditioned on actions, enabling planning and sample-efficient
policy learning. However, RL world models assume dense regular observations, closed-loop
action selection, and episodic resets. Industrial open-loop simulation instead works with
sparse irregular sensing, externally specified future control sequences, and continuous
operation. CCSS-RS adapts the world-model idea to this setting by prioritising
long-horizon stability, missingness handling, and calibrated rollout uncertainty over
exploration and value estimation.

\subsection{Probabilistic Forecasting and WWTP Data-Driven Modelling}

For probabilistic output, DeepAR \citep{salinas2020deepar} and DeepState
\citep{rangapuram2018deep} pioneer likelihood-based training with autoregressive RNNs
and state-space models respectively, while diffusion-based methods \citep{rasul2021timegrad,tashiro2021csdi}
offer flexible multimodal distributions at greater computational expense. In the Aved{\o}re dataset used here, WWTP sensor distributions exhibit kurtosis up to 331.9
and zero-inflation exceeding 35\%, making Gaussian likelihoods systematically miscalibrated;
Student-$t$ likelihoods are therefore appropriate for the heavy-tailed marginals, while
hurdle mechanisms \citep{mullahy1986specification} address the structural zero mass.

The ASM family \citep{henze2000asm} of mechanistic WWTP models encodes decades of
biochemical domain knowledge but suffers from identifiability issues, a large number of parameters requiring expert
calibration, and systematic mismatch between model assumptions and real plant configurations
\citep{gernaey2004activated}. Data-driven approaches are gaining traction for WWTP soft-sensing and performance monitoring
\citep{newhart2019data}, yet remain focused on single-step prediction, relatively short
horizons, or offline analytics. The Aved{\o}re dataset \citep{hansen2024avedore} was
released specifically to enable ML research on full-scale WWTP data with realistic
irregularity; long-horizon fidelity beyond a few hundred steps under this data regime
remains the open challenge that CCSS-RS addresses.

\subsection{Baseline Selection}
\label{sec:baseline_selection}

Industrial WWTP data presents five requirements that must be simultaneously satisfied for
fair baseline comparison: native irregular sampling without resampling (R1,
$\Delta t \in [1,20]$~min); missing data handling without imputation (R2, 43\%
missingness); leverage of known future controls and exogenous inputs (R3); categorical
variable embedding (R4); and stable rollout to extreme long horizons (R5, $H=1000$).
Table~\ref{tab:baseline_survey} summarises how major model families align with these
requirements.

\textbf{Transformer-based models} (TFT, Informer, PatchTST) do not natively satisfy R1
and R2 under the no-resampling protocol because standard implementations operate on
regular token grids, while forward-filling 43\% missingness over 1000 steps would
introduce false regularity and incomparable evaluation conditions.
\textbf{SSMs} (S4, Mamba, MambaTS) face the same limitation in their standard form:
they assume a fixed discrete grid, and adapting them to arbitrary timestamps with 43\%
missingness would require additional modelling choices that are not part of the native
baseline.
\textbf{RNN variants} such as GRU-D partially address R1 via learned exponential decay,
but cannot leverage known future controls (R3) without either bidirectional processing
(which breaks causality) or a regular-grid decoder; gradient degradation also limits
reliable rollout beyond 200--500 steps (R5). \textbf{Linear models} (DLinear, TiDE)
operate on fixed-length regular windows and require resampling or a redesign for irregular timestamps (R1).
\textbf{Neural ODEs} naturally handle R1 by integrating at arbitrary evaluation times,
but the standard autonomous formulation $dz/dt = f_\theta(z,t)$ has no native mechanism
for externally prescribed control paths (R3), and sparse context makes initial-state
inference difficult at this missingness level.

\textbf{Neural CDEs} \citep{kidger2020neural} come closest to satisfying all five
requirements under the no-resampling protocol. Their continuous-time formulation is
intrinsically irregular-sampling aware (R1); mask-aware path construction can accommodate
missing values without forward-fill (R2); the control path can include specified future
controls and exogenous forecasts (R3); categorical variables can be embedded alongside
continuous signals (R4); and solver-controlled rollouts can be evaluated over long horizons
(R5). Neural CDEs are therefore the most defensible external baseline, included at two
capacity levels (0.67M and 4.1M parameters) to control for model-size effects. Controlled
ablation of CCSS-RS components (Section~\ref{sec:ablation}) then answers the
complementary question: given that this data regime must be handled, which architectural
choices most influence long-horizon performance?

\begin{table*}[t]
\centering
\caption{Baseline compatibility with WWTP simulation requirements.
\textbf{Legend:} \cmark~=~native support; \trimark~=~partial / workaround required;
\xmark~=~not natively supported under this protocol.}
\label{tab:baseline_survey}
\small
\resizebox{\textwidth}{!}{%
\begin{tabular}{llcccccl}
\toprule
\textbf{Model Family} & \textbf{Representative Models}
  & \textbf{R1} & \textbf{R2} & \textbf{R3} & \textbf{R4} & \textbf{R5}
  & \textbf{Comparable?} \\
  & & \textbf{Irregular} & \textbf{No Impute} & \textbf{Controls}
  & \textbf{Categorical} & \textbf{Long Horizon} & \\
\midrule
Transformers           & TFT, Informer, FEDformer, PatchTST, Autoformer
  & \xmark & \xmark & \cmark & \cmark & \trimark & No \\
State Space (S4/Mamba) & S4, Mamba, MambaTS, S5, GPS-Mamba
  & \xmark & \xmark & \trimark & \cmark & \cmark & No \\
RNN Variants           & LSTM, GRU-D, DeepAR, N-BEATS
  & \trimark & \trimark & \trimark & \cmark & \xmark & No \\
Linear Models          & DLinear, TiDE, TSMixer
  & \xmark & \xmark & \trimark & \trimark & \cmark & No \\
Neural ODEs            & Neural ODE, Latent ODE
  & \cmark & \trimark & \xmark & \cmark & \cmark & No \\
Neural CDEs            & Neural CDE, Log-NCDE, Stable Neural SDE
  & \cmark & \cmark & \cmark & \cmark & \cmark & \textbf{Yes} \\
\bottomrule
\end{tabular}}
\end{table*}

\section{Problem Formulation}
\label{sec:problem}

Consider a controlled industrial process with three variable types: \textbf{state
variables} $x(t) \in \mathbb{R}^{D_x}$ (measurable outputs we predict: $\text{NH}_4$,
$\text{NO}_3$, $\text{N}_2\text{O}$, $\text{O}_2$, SS); \textbf{control inputs}
$u(t) \in \mathbb{R}^{D_u}$ (operator-specified actions: valve positions, aeration
rates); and \textbf{exogenous disturbances} $w(t) \in \mathbb{R}^{D_w}$ (external
influences: influent flow, temperature). True system dynamics follow an unknown function
$dx/dt = f_{\text{true}}(x(t), u(t), w(t), t)$. We observe the system through sensors
sampling irregularly at times $\{t_1,\ldots,t_N\}$ with variable gaps
$\Delta t_n = t_n - t_{n-1}$, receiving at $t_n$ a partial observation
$o_n = \{x^{(i)}(t_n) : i \in \mathcal{O}_n\}$ where
$\mathcal{O}_n \subseteq \{1,\ldots,D_x\}$.

Given historical context
$\{o_1,\ldots,o_T\}$,
$\{u_1,\ldots,u_T\}$, and
$\{w_1,\ldots,w_T\}$
over $[t_1,t_T]$ together with a known future plan
$\{u_{T+1},\ldots,u_{T+H}\}$
and exogenous forecast
$\{w_{T+1},\ldots,w_{T+H}\}$,
the open-loop simulation task is to produce:
\begin{equation}
    p_\theta\!\left(x_{T+1:T+H} \mid \text{context},\, u_{T+1:T+H},\, w_{T+1:T+H}\right)
\end{equation}
This differs from standard forecasting in that all future inputs are \emph{known}, long horizons of $H \approx 500$--$1000$ steps are used in our MPC-style screening setting, and predictions must
remain numerically stable without diverging. We evaluate using five complementary metrics:
RMSE and MAE for point accuracy, CRPS as a proper scoring rule for probabilistic quality,
NLL for distributional calibration, and DTW for alignment-aware evaluation.

\section{The CCSS Architecture Family}
\label{sec:architecture}

\subsection{Design Rationale}
\label{sec:rationale}

Four industrial requirements shape CCSS-RS directly, each addressed by a corresponding
architectural choice (Table~\ref{tab:design_rationale}).

\begin{table}[h]
\centering
\caption{Industrial requirements and their CCSS-RS architectural responses.}
\label{tab:design_rationale}
\small
\setlength{\tabcolsep}{4pt}
\begin{tabular}{>{\raggedright\arraybackslash}p{3.6cm} >{\raggedright\arraybackslash}p{4.6cm}}
\toprule
\textbf{Industrial requirement} & \textbf{Architectural response} \\
\midrule
Future control plans are fully known; process states must be inferred
  & Hard typed-stream separation: controls route \emph{only} into the dynamics function;
    historical observations initialise the latent belief state once \\
\addlinespace
Sensing is irregular (1--20\,min intervals) with 43\% missingness; no resampling
  & Continuous-time affine scans with $\Delta t$-aware exponential decay; five-feature
    per-variable encoding including observation mask and log-age-since-last-observation \\
\addlinespace
WWTP dynamics switch abruptly between aerobic/anoxic phases; regime errors compound
  & Sticky HDP-HMM-inspired router over $K=3$ regime-specific experts; a temporal
    persistence prior prevents unrealistic rapid switching \\
\addlinespace
Long rollouts ($H=1000$) accumulate discretisation error; sensor distributions are
    heavy-tailed and zero-inflated
  & Semigroup-consistency-inspired regularisation \citep{chen2023deeposg}; Student-$t$ likelihood
    with hurdle mechanism for $\text{N}_2\text{O}$ \\
\bottomrule
\end{tabular}
\end{table}

Neural CDEs \citep{kidger2020neural} are the closest external family because they natively
handle irregular sampling, but they typically encode past observations, future controls,
and exogenous forecasts within a single driving path, rather than making the control-state
distinction explicit in the model structure. CCSS-RS encodes this distinction as a hard
architectural prior: historical observations initialise the latent belief state, while
future controls enter only through the rollout dynamics that simulate the process state.

\begin{figure*}[t]
    \centering
    \includegraphics[width=0.99\linewidth]{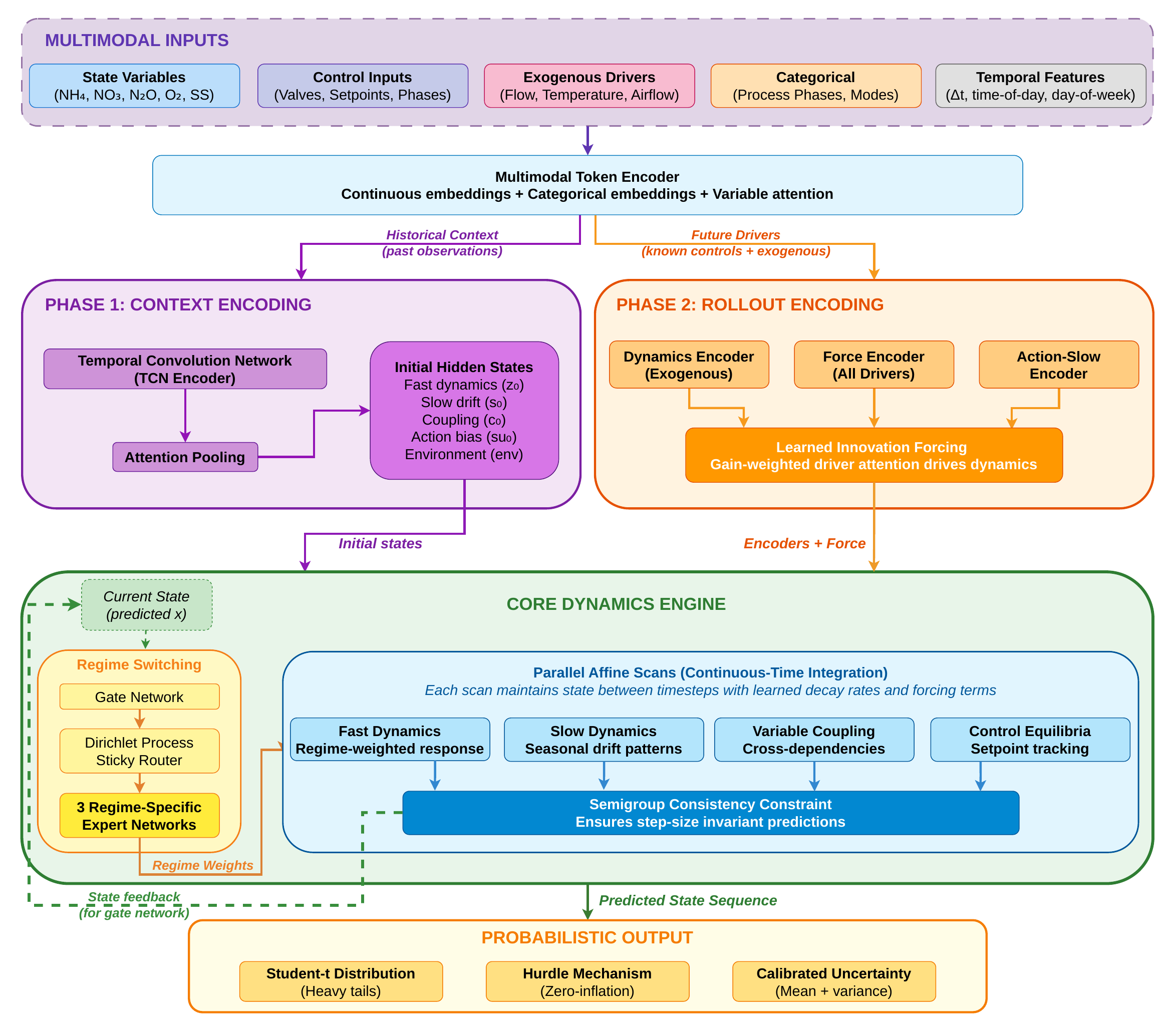}
    \caption{CCSS-RS architecture. \textbf{Phase~1} (left): historical context is encoded
    by a TCN and attention pooling into a context summary $\text{env}$ and four typed
    initial hidden states ($z_0$, $s_0$, $c_0$, $su_0$). \textbf{Phase~2} (right): three
    parallel encoders process prescribed controls and forecast exogenous drivers; the core dynamics engine integrates the
    state trajectory via four parallel affine scans modulated by regime-specific experts;
    a learned gain-weighted forcing mechanism derived from future driver tokens directly
    drives fast-dynamics evolution; and a semigroup consistency constraint enforces
    step-size invariance. Outputs are probabilistic, using Student-$t$ emissions with a
    hurdle mechanism for zero-inflated variables.}
    \label{fig:architecture}
\end{figure*}

The model operates in two sequential phases (Figure~\ref{fig:architecture}).
\textbf{Phase~1} (context encoding) reads historical plant data over a 512-step window
and distils it into a typed belief state. \textbf{Phase~2} (rollout) advances the
learned dynamics forward under the specified future control plan, producing probabilistic
state trajectories.

\subsection{Architecture}
\label{sec:arch_detail}

\paragraph{Input representation.}
All signals pass through a shared multimodal token encoder. Continuous measurements are
standardised and projected with extreme-value clipping at $|z|>25$; categorical variables
use 32-dimensional lookup tables with mask-aware, age-decayed interpolation; temporal
features include $\Delta t$, time-of-day, and day-of-week. Each variable carries five
features: observed value, last-observation belief, observation mask, log-age-since-last
observation, and rate-of-change. State variables, controls, and exogenous drivers are
kept in separate typed streams throughout both phases.

\paragraph{Phase 1: context encoding.}
A TCN \citep{bai2018empirical} with four dilated depthwise-separable convolution blocks
and GLU activations encodes the 512-step historical window, providing a multi-scale context
representation from short-term fluctuations to longer trends. Attention pooling then compresses the sequence
into a context summary $\text{env}$ and four typed initial hidden states that mirror the
physical timescale structure of WWTP operation: \textbf{fast dynamics} $z_0
\in \mathbb{R}^{256}$ (rapid state responses to control actions); \textbf{slow drift}
$s_0 \in \mathbb{R}^{16}$ (multi-hour biological accumulation); \textbf{variable
coupling} $c_0 \in \mathbb{R}^{64}$ (cross-state dependencies, e.g.\ the
$\text{NH}_4$--$\text{NO}_3$ balance during nitrification); and \textbf{control
equilibria} $su_0 \in \mathbb{R}^{16}$ (the learned state offset from prior control
history). The simplified variants CCSS-SR and CCSS-G replace the TCN with Selective IIR encoder
stacks \citep{orvieto2023resurrecting}. These converge faster and require less memory,
but are less effective under the sharp regime transitions observed in this WWTP setting
(see Section~\ref{sec:results}).

\paragraph{Phase 2: rollout encoders.}
Three specialised TCN encoders process prescribed controls and forecast exogenous drivers in parallel at each
rollout step. The \textbf{Dynamics Encoder} ($\text{h}_{\text{dyn}}$) processes
exogenous variables alone (influent flow, temperature), capturing how ambient disturbances
shape state evolution independently of operator decisions. The \textbf{Force Encoder} ($\text{h}_{\text{force}}$) processes all future drivers
jointly to produce a global forcing representation of their combined influence on the
system dynamics. The \textbf{Action-Slow Encoder} ($\text{h}_{\text{actslow}}$) focuses
on control inputs, learning how their effects propagate gradually into process states over
hours, for example through the delayed impact of aeration setpoint changes on ammonia
concentrations. All three feed the core dynamics engine at every step.

\paragraph{Gain-weighted innovation forcing.}
Encoder conditioning alone is insufficient to capture control responsiveness; driver
signals must also be injected directly into the evolving state trajectory. At each step, the Force
Encoder produces per-variable token representations; a token-level gain network computes
attention weights $g_{t,k} = \sigma(f_{\text{gain}}([{\rm tok}_k, \bar{\rm tok}]))$ and
assembles the forcing signal as:
\begin{equation}
    F_t = \sum_{k \in \mathcal{A}} g_{t,k}\, f_{\text{act}}({\rm tok}_k) +
          \sum_{k \in \mathcal{E}} g_{t,k}\, f_{\text{exo}}({\rm tok}_k)
\end{equation}
where $\mathcal{A}$ and $\mathcal{E}$ index action and exogenous variables. $F_t \in
\mathbb{R}^{Z}$ is injected directly into the fast-dynamics scan at each step. An
auxiliary head (weight 0.20) predicts observed control values from the hidden state,
providing a training signal that sharpens control encoding. Ablation confirms removing
this mechanism degrades RMSE by 11.0\% at $H=1000$ (Section~\ref{sec:ablation}).

\paragraph{Four parallel affine scans.}
The core engine integrates the four typed hidden states via \emph{parallel affine scans}
\citep{smith2023s5}: $h_t = a_t h_{t-1} + b_t$, where $a_t = \exp(-\lambda\,\Delta
t_{\text{eff}})$ enforces exponential decay with natural $\Delta t$-awareness. The
\textbf{fast scan} ($z$, dim~256) integrates rapid responses to $F_t$ and slow-scan
output, with regime-specific decay rates $\lambda_z^{(k)}$. The \textbf{slow scan} ($s$,
dim~16) integrates biological accumulation and couples upward to the fast scan. The
\textbf{coupling scan} ($c$, dim~64) maintains cross-state dependency information. The
\textbf{equilibria scan} ($su$, dim~16) tracks the state offset from accumulated control
history, contributing an additive bias to emission means. Integration runs for two Picard
iterations: the first initialises from prior belief $x_0$; the second uses $\hat{x}$ from
the first pass for more accurate regime routing.

\paragraph{Sticky regime switching.}
\label{sec:regime_switch}
Control variables alone do not fully characterise the operational regime: biological
processes exhibit hysteresis, and aeration phase dynamics depend on the accumulation
history that precedes a transition. A gate network maps the predicted state sequence and
$\text{h}_{\text{dyn}}$ to soft weights over $K=3$ regime-specific expert networks via a sticky HDP-HMM-inspired router
\citep{teh2006hierarchical,fox2011sticky}. Sticky transitions
$p(z_t|z_{t-1}) = \rho\cdot\delta(z_t=z_{t-1}) + (1-\rho)\cdot p(z_t)$
enforce temporal persistence (initial $\rho=0.95$). Expert contributions combine as:
\begin{equation}
    \frac{dx}{dt} = \sum_k p(z=k|\text{history}) \cdot f_{\theta_k}(x, u, w, b)
\end{equation}
Removing regime switching degrades RMSE by 5.2\% at $H=1000$, with the gap growing
super-linearly as mode errors cascade across subsequent phase transitions.

\paragraph{Semigroup consistency.}
\label{sec:semigroup}
True differential equations satisfy the semigroup property: evolving from $0$ to
$t_1 + t_2$ produces the same result as evolving to $t_1$ and continuing to $t_1 + t_2$.
Discretised learned dynamics can violate this property, leading to step-size-dependent drift at
long horizons; we adopt a semigroup-consistency regulariser in the spirit of
\citet{chen2023deeposg}. Denoting the learned evolution operator
$\Phi_\theta(x, t_1{\to}t_2)$, CCSS-RS enforces:
\begin{equation}
    \Phi_\theta(x, 0 \to t_1{+}t_2) = \Phi_\theta\!\left(\Phi_\theta(x, 0 \to t_1),\, t_1 \to t_1{+}t_2\right)
\end{equation}
stochastically (prob.\ 0.10 per batch, weight 0.20) by penalising the squared mean-state
discrepancy between the full-horizon and composed two-segment predictions. Removing this
constraint degrades RMSE by 11.7\%, with the gap growing linearly with horizon.

\paragraph{Probabilistic output.}
In the Aved{\o}re dataset used here, all five state variables exhibit kurtosis $>3$ (up to 331.9 for $\text{N}_2\text{O}$),
making Gaussian likelihoods systematically miscalibrated. CCSS-RS uses a Student-$t$
likelihood with learned per-state degrees-of-freedom $\nu$ in log$1p$-transformed space,
plus a tail loss ($3\times$ weight above the 95th-percentile) to prevent collapse on
heavy-tailed events. For $\text{N}_2\text{O}$ (35\% structural zeros), a hurdle mechanism
separates the zero/non-zero decision:
\begin{equation}
    p(x) = \begin{cases}
        \pi & x = 0 \\
        (1-\pi)\cdot\mathrm{Student\text{-}}t(x;\,\mu,\sigma,\nu) & x > 0
    \end{cases}
\end{equation}
where $\pi$ evolves via its own affine scan. Replacing this with a Gaussian likelihood
(CCSS-G) increases RMSE by 53\% and NLL by an order of magnitude at $H=1000$, confirming
the heavy-tailed design is a data necessity rather than a modelling preference.

\subsection{Model Variants and Training}
\label{sec:variants}

Table~\ref{tab:variants} summarises the model family. CCSS-RS (6.9M) uses TCN-based
encoders for both phases; simplified variants CCSS-SR and CCSS-G use Selective IIR encoder
stacks inspired by recent long-sequence recurrent designs \citep{orvieto2023resurrecting}.
In our experiments, these variants are lighter-weight and converge faster, but are less
suited to abrupt regime transitions. Despite having more parameters overall (7.3M), CCSS-SR and CCSS-G perform substantially
worse. This confirms that inductive biases, including typed hidden states, the
three-encoder rollout structure, and regime switching, matter more than raw capacity in
this setting. Neural CDE baselines provide an external reference at two capacity levels.

\begin{table*}[t]
\centering
\caption{Model variant comparison including computational efficiency on a single NVIDIA
RTX PRO 6000 Blackwell GPU. Train time is per full 25-epoch run; inference is per
1000-step rollout window on CPU. \cmark~=~enabled; N/A~=~not applicable.}
\label{tab:variants}
\small
\setlength{\tabcolsep}{8pt}
\begin{tabular}{llcclcc}
\toprule
\textbf{Model} & \textbf{Params} & \textbf{Train (min)} & \textbf{Infer (s)} & \textbf{Regime Switching} & \textbf{Likelihood} & \textbf{Innovation Forcing} \\
\midrule
CCSS-RS (ours) & 6.9M  & \textbf{6.5}  & \textbf{1.2} & DP Mixture & Student-$t$ + Hurdle & \cmark \\
CCSS-SR (ours) & 7.3M  & 5.9           & 0.8          & None       & Student-$t$ + Hurdle & \cmark \\
CCSS-G (ours)  & 7.3M  & 5.8           & 0.8          & None       & Gaussian             & \cmark \\
NCDE-Large     & 4.1M  & 65.4          & 4.8          & N/A        & Gaussian             & N/A   \\
NCDE-Small     & 0.67M & 9.8           & 2.5          & N/A        & Gaussian             & N/A   \\
\bottomrule
\end{tabular}
\end{table*}

Training uses a composite objective: primary NLL with tail weighting (weight 1.0); hurdle
binary cross-entropy for $\text{N}_2\text{O}$ zero classification (0.60); semigroup
consistency penalty (0.20); innovation alignment ensuring $F_t$ amplifies information not
captured by base scan decay (0.18); regime regularisation combining total variation,
KL-divergence usage penalty, DP stick-breaking prior, and entropy terms; action auxiliary
loss predicting observed control values from hidden state (0.20); and temporal smoothness
on second-order trajectory differences ($2\!\times\!10^{-3}$). A time-weighted loss
$w(t) = 0.2 + 2.0\cdot(t/H)^2$ emphasises long-horizon accuracy. Training uses AdamW
($\text{lr}=8\!\times\!10^{-4}$, weight decay $0.01$), batch size 40, for 25 epochs.
Notably, CCSS-RS trains 10$\times$ faster than NCDE-Large (6.5 vs.\ 65.4 minutes) and
runs inference 4$\times$ faster (1.2 vs.\ 4.8 seconds per window), while achieving
40--46\% lower RMSE, indicating that the architectural inductive biases improve
accuracy without sacrificing computational efficiency.

\section{Experimental Setup}
\label{sec:experimental}

We use the Aved{\o}re WWTP dataset \citep{hansen2024avedore}: 906,815 timesteps over two
years (June 2022--June 2024) at a municipal facility serving 350,000 population
equivalent near Copenhagen, Denmark. Biological treatment in four parallel lines uses an
alternating activated sludge process (ASP) regulated by ammonium, nitrate, and dissolved
oxygen concentrations. We use 16 variables from Tank 1 organised into state variables
($\text{NH}_4$, $\text{NO}_3$, $\text{N}_2\text{O}$, $\text{O}_2$, SS), control
variables (PCT valve positions, oxygen setpoints, phase indicators), exogenous variables
(influent flow, blower airflow, temperature), and categorical variables (process phases).

The dataset is challenging because it combines properties rarely present together in
standard benchmarks: 43.4\% overall missingness (57.4\% for $\text{N}_2\text{O}$;
consecutive PCT gaps up to 21 days); informative missingness correlated with process
phases; sampling intervals from 1 to 20 minutes precluding fixed grids; all five state
variables failing normality tests with kurtosis 6.7--331.9; 35\% zero-inflation in
$\text{N}_2\text{O}$; discrete phase transitions every 2--3 hours; and evaluation at $H=1000$ steps, far
exceeding the $H=96$ typical of forecasting benchmarks and matching the long planning
horizons used in our MPC-style screening setting. These properties make Aved{\o}re
substantially more challenging than standard forecasting benchmarks and unusual among
real-world datasets used for long-horizon simulation.

We use a temporal train/test split of 70\%/30\%, with 11,839 eligible test windows
(4.4\% of 270,532 candidates) defined by requiring all 1000 rollout timesteps to be
fully observed. For the main model comparison, 10,000 windows are randomly sampled
(seed=42). For horizon sweeps and ablations, 5,000 windows are evaluated for horizons
$H \in [100,1000]$ in steps of 100, with 95\% bootstrap confidence intervals
(3,000 resamples). Neural CDE baselines are excluded from ablation analyses, which
isolate contributions within the CCSS family.

\section{Results}
\label{sec:results}

Before reporting metrics, it is worth anchoring the numbers in operating context. At
Aved{\o}re, $\text{NH}_4$ spans 0.1--12\,mg-N/L during normal operation and is the
primary effluent compliance target; $\text{NO}_3$ spans 0.5--10\,mg-N/L and reflects
denitrification efficiency; $\text{O}_2$ is tightly controlled at 0.5--2.0\,mg/L by the
aeration system; $\text{N}_2\text{O}$ is a greenhouse-gas emission indicator with a
wide dynamic range (0--2\,mg-N/L) and heavy right tail; and SS tracks suspended-solids
loading. An RMSE of 0.8\,mg-N/L for $\text{NH}_4$ at $H=1000$ represents uncertainty of
roughly $\pm$0.8\,mg-N/L on a 12\,mg-N/L dynamic range. This is practically useful
for distinguishing materially different future scenarios at the 12--36\,h planning
horizon of practical interest. The purpose of the model is relative plan comparison, not
exact trajectory prediction, so the relevant question is whether the simulator produces
informative differences among candidate plans rather than whether every predicted value is
within a tight absolute tolerance. We return to
this point in the case studies (Section~\ref{sec:case_studies}).

\subsection{Aggregate Performance at \texorpdfstring{$H=1000$}{H=1000}}

We first report aggregate performance at $H=1000$, then examine per-variable behaviour,
difficulty stratification, and the horizon profile before turning to application-oriented
case studies.

Table~\ref{tab:main_results} summarises performance across all five models. CCSS-RS
achieves the best results on all four metrics. The point-error ordering is consistent:
CCSS-RS $\succ$ CCSS-SR $\succ$ CCSS-G $\succ$ NCDE-Large $\succ$ NCDE-Small. The NLL
ordering differs: CCSS-G collapses to NLL of 6.65 because the Gaussian likelihood
fundamentally cannot model heavy-tailed and zero-inflated distributions, while CCSS-SR
and the Neural CDEs perform comparably (1.51, 1.08, 1.12). Interestingly, CCSS-G ranks
second on CRPS (0.523) through a degenerate strategy of predicting near-zero
$\text{N}_2\text{O}$ constantly, which exploits the 35\% zero mass but is exposed by its
catastrophic NLL.

\begin{table}[h]
\centering
\caption{Model performance at $H=1000$ over 10,000 evaluation windows (lower is better).}
\label{tab:main_results}
\small
\resizebox{\columnwidth}{!}{%
\begin{tabular}{lccccc}
\toprule
\textbf{Model} & \textbf{Params (M)} & \textbf{MAE} & \textbf{RMSE} & \textbf{NLL} & \textbf{CRPS} \\
\midrule
CCSS-RS    & 6.9  & \textbf{0.452} & \textbf{0.696} & \textbf{0.585} & \textbf{0.349} \\
CCSS-SR    & 7.3  & 0.649          & 1.003          & 1.514          & 0.534          \\
CCSS-G     & 7.3  & 0.660          & 1.066          & 6.651          & 0.523          \\
NCDE-Small & 0.7  & 0.800          & 1.279          & 1.076          & 0.573          \\
NCDE-Large & 4.1  & 0.734          & 1.155          & 1.118          & 0.528          \\
\bottomrule
\end{tabular}}
\end{table}

The gap between CCSS-RS and simplified variants is substantial: CCSS-SR is 44\% worse
on RMSE (1.003 vs.\ 0.696) despite having more parameters; CCSS-G is 53\% worse. Against
Neural CDE baselines, CCSS-RS achieves 40--46\% lower RMSE, 34--39\% lower CRPS, and
46--48\% lower NLL. Scaling Neural CDE from 0.7M to 4.1M yields only modest
improvement (RMSE 1.279 $\to$ 1.155), confirming the performance gap is architectural
rather than capacity-driven.

\subsection{Per-Variable and Difficulty-Stratified Analysis}

Table~\ref{tab:per_state} breaks down performance by state variable. $\text{O}_2$ is the
easiest to predict (CCSS-RS RMSE 0.273), reflecting tight control by the aeration system;
CCSS-RS wins in over 91\% of evaluation windows. $\text{NO}_3$ is the hardest, with RMSE
above 1.6 for all models except CCSS-RS, consistent with complex nutrient cycling and
strong coupling to the aeration regime. The $\text{N}_2\text{O}$ results require care:
CCSS-G achieves the lowest point-error (MAE 0.075, RMSE 0.102) by predicting near-zero
constantly, which yields low MAE on the 35\% zero mass but fails on NLL (13.7 vs.\ 0.47
for CCSS-RS), confirming that CCSS-RS and CCSS-SR correctly model zero-inflated dynamics
via the hurdle mechanism while CCSS-G does not.

\begin{table}[h]
\centering
\caption{Per-state variable performance at $H=1000$ (mean MAE / RMSE over 10,000 windows).}
\label{tab:per_state}
\small
\setlength{\tabcolsep}{4pt}
\begin{tabular}{lcccccc}
\toprule
 & \multicolumn{2}{c}{\textbf{CCSS-RS}} & \multicolumn{2}{c}{\textbf{CCSS-SR}} & \multicolumn{2}{c}{\textbf{CCSS-G}} \\
\cmidrule(lr){2-3}\cmidrule(lr){4-5}\cmidrule(lr){6-7}
\textbf{Variable} & MAE & RMSE & MAE & RMSE & MAE & RMSE \\
\midrule
$\text{NH}_4$ & \textbf{0.659} & \textbf{0.822} & 0.833 & 1.026 & 1.129 & 1.339 \\
$\text{NO}_3$ & \textbf{0.842} & \textbf{0.998} & 1.484 & 1.701 & 1.409 & 1.639 \\
$\text{N}_2\text{O}$ & 0.107 & 0.143 & 0.138 & 0.177 & \textbf{0.075} & \textbf{0.102} \\
$\text{O}_2$ & \textbf{0.185} & \textbf{0.273} & 0.231 & 0.347 & 0.225 & 0.330 \\
SS            & 0.465 & \textbf{0.538} & 0.558 & 0.644 & \textbf{0.459} & 0.545 \\
\bottomrule
\end{tabular}
\end{table}

Stratifying the 5,000 ablation windows by a composite difficulty score (z-score sum of
persistence-baseline RMSE, within-window standard deviation, and state excursion range),
CCSS-RS leads consistently in both high- and low-difficulty subsets
(Table~\ref{tab:hard_easy}). On high-difficulty windows the RMSE gap over CCSS-SR is
42\% (0.728 vs.\ 1.032), reflecting the value of regime switching during sharp
transitions. The relative advantage is equally large on easy windows (37\% gap),
indicating that regime switching and innovation forcing improve trajectory tracking even
during nominally stable periods, since ``easy'' windows still contain subtle mode changes
that global smooth models handle less well.

\begin{table}[t]
\centering
\caption{Performance on higher- vs.\ lower-difficulty windows at $H=1000$
(mean $\pm$ std; lower is better).}
\label{tab:hard_easy}
\small
\setlength{\tabcolsep}{5pt}
\begin{tabular}{@{}llcc@{}}
\toprule
\textbf{Subset} & \textbf{Model} & \textbf{RMSE} & \textbf{CRPS} \\
\midrule
\multirow{3}{*}{Higher difficulty} & CCSS-RS & $\mathbf{0.728 \pm 0.117}$ & $\mathbf{0.379 \pm 0.064}$ \\
 & CCSS-SR & $1.032 \pm 0.204$ & $0.557 \pm 0.113$ \\
 & CCSS-G  & $1.055 \pm 0.212$ & $0.529 \pm 0.121$ \\
\midrule
\multirow{3}{*}{Lower difficulty} & CCSS-RS & $\mathbf{0.546 \pm 0.254}$ & $\mathbf{0.295 \pm 0.144}$ \\
 & CCSS-SR & $0.865 \pm 0.277$ & $0.476 \pm 0.157$ \\
 & CCSS-G  & $0.846 \pm 0.445$ & $0.446 \pm 0.250$ \\
\bottomrule
\end{tabular}
\end{table}

\begin{figure}
\centering
\includegraphics[width=0.99\linewidth]{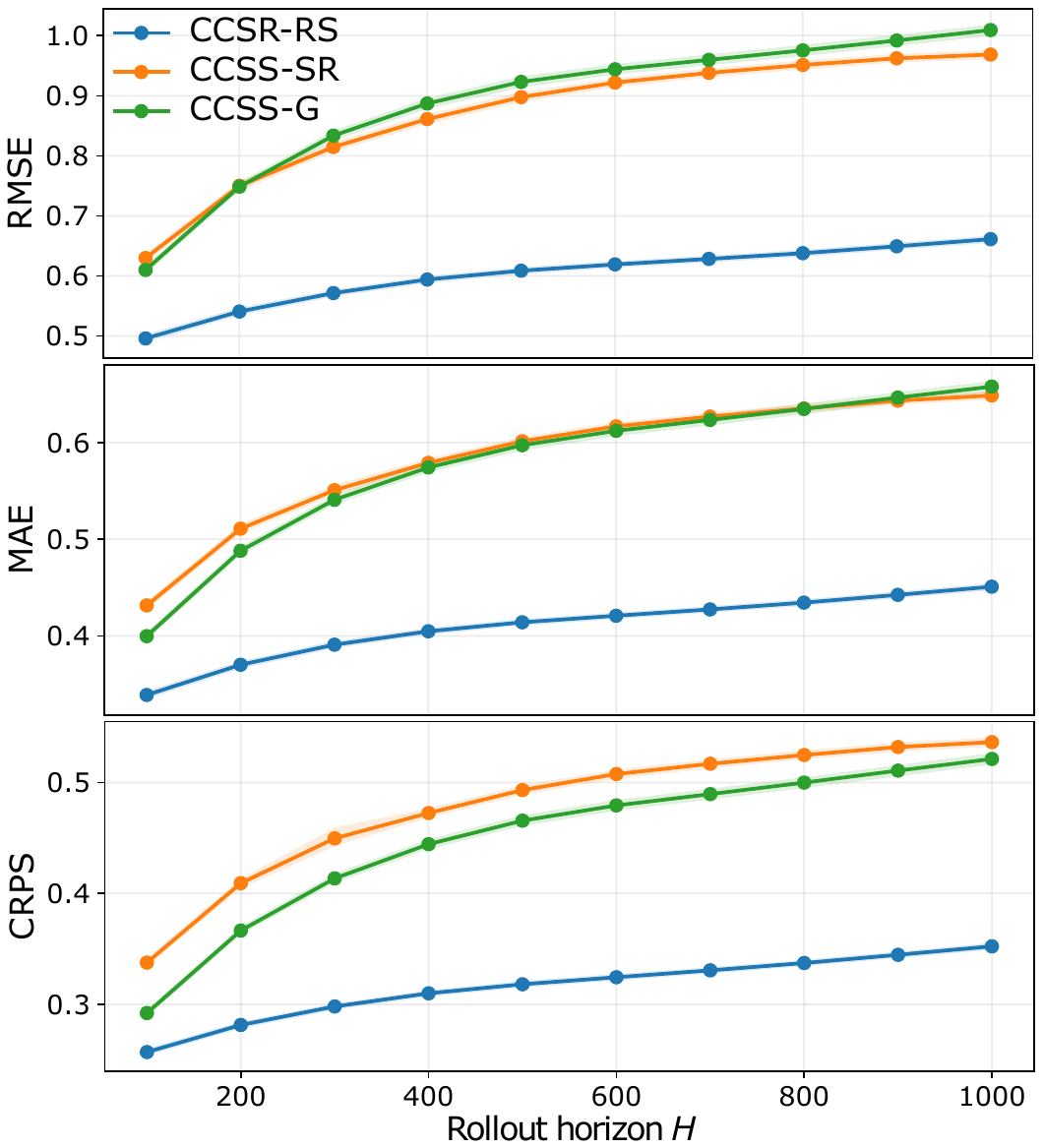}
\caption{RMSE (top), MAE (middle), and CRPS (bottom) vs.\ rollout horizon $H$ on 5,000
fully-observed windows. Bootstrap 95\% confidence intervals (3,000 resamples) are plotted
but negligibly narrow at this scale.}
\label{fig:horizon_sweep}
\end{figure}

\subsection{Horizon Sweep and Qualitative Analysis}

Figure~\ref{fig:horizon_sweep} reports RMSE, MAE, and CRPS as the rollout horizon
extends from $H{=}100$ to $H{=}1000$ on the 5,000-window horizon-sweep subset.
CCSS-RS accumulates error substantially more slowly: RMSE grows from 0.496 to 0.661 (+33\%), compared to 0.630 $\to$ 0.969 for
CCSS-SR (+54\%) and 0.610 $\to$ 1.009 for CCSS-G (+65\%). The gap between CCSS-RS and CCSS-SR is modest at short horizons (2.9\% at $H{=}100$)
but grows to 9.3\% by $H{=}1000$, consistent with architectural choices for regime
handling and rollout conditioning becoming more important as mode errors compound through
subsequent ASP phases. CRPS grows more slowly than RMSE across all models,
consistent with predictive uncertainty broadening to partially compensate for increased
trajectory error.

\begin{figure*}[t]
\centering
\includegraphics[width=1.0\linewidth]{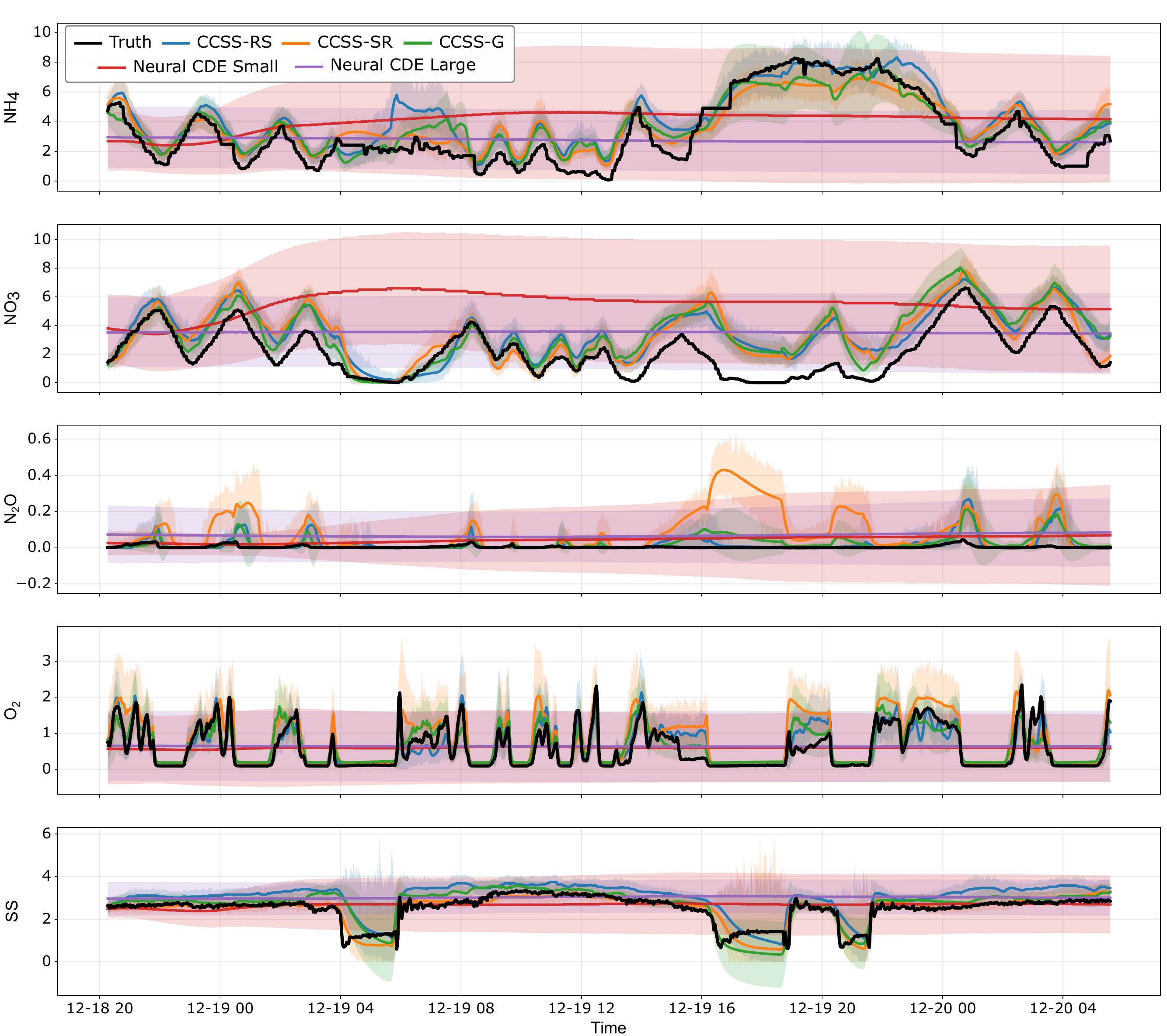}
\caption{Prediction trajectories for Window~1 (December 18--20), characterised by rapid
$\text{NH}_4$ oscillations, sharp SS drops, and near-zero $\text{N}_2\text{O}$. Shaded
regions: 95\% predictive intervals.}
\label{fig:window1}
\end{figure*}

Qualitatively, Window~1 (December 18--20) presents regular $\text{O}_2$ oscillations
driven by aeration cycles, rapid $\text{NH}_4$ fluctuations from alternating
nitrification phases, and sharp SS drops coinciding with settling. CCSS-RS closely
tracks the $\text{O}_2$ cycle amplitude and period throughout 32 hours, captures the
$\text{NH}_4$ surge in the second half with appropriate lag, and produces predictive
intervals that tighten during stable aeration phases and widen around phase transitions.
CCSS-SR shows broadly similar cycle tracking but with increased lag on $\text{NH}_4$
peaks, consistent with reduced sensitivity to mode changes. CCSS-G tracks the
$\text{NH}_4$ and $\text{NO}_3$ mean levels but drifts progressively upward, and its
$\text{N}_2\text{O}$ predictions collapse to near-zero without meaningful uncertainty.
Both Neural CDEs produce near-constant trajectories for all five state variables,
failing to represent any cyclic or trend dynamics (Figure~\ref{fig:window1}).

\begin{figure*}[t]
\centering
\includegraphics[width=1.0\linewidth]{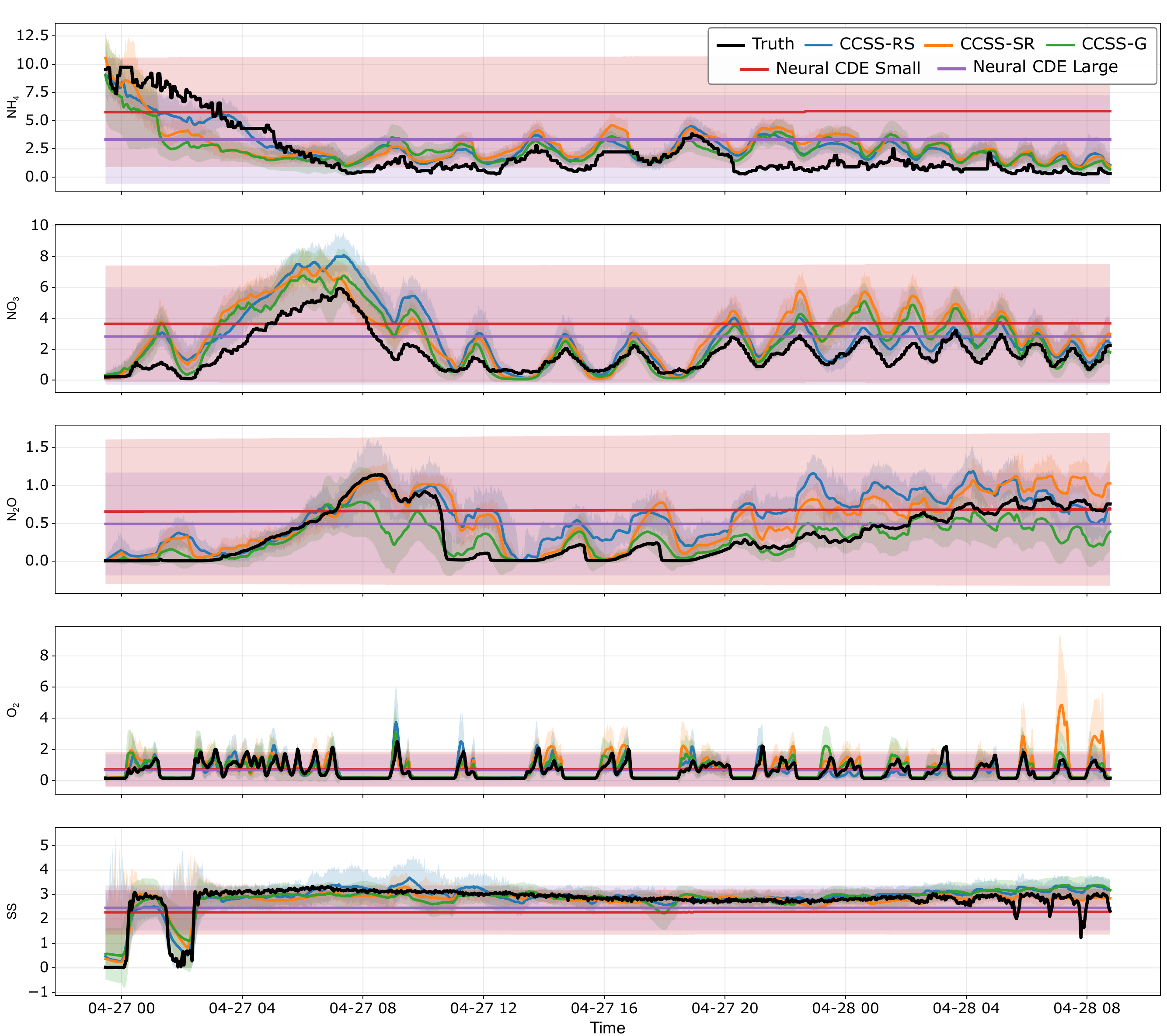}
\caption{Prediction trajectories for Window~2 (April 27--28), featuring a pronounced
$\text{NH}_4$ decay from $\sim$10 to $\sim$2, rising $\text{N}_2\text{O}$, and sustained
$\text{O}_2$ oscillations. Shaded regions: 95\% predictive intervals.}
\label{fig:window2}
\end{figure*}

Window~2 (April 27--28) highlights a different failure mode: a long-range transient
where $\text{NH}_4$ decays from $\sim$10 to $\sim$2 over six hours as nitrification
intensifies, accompanied by a corresponding $\text{NO}_3$ rise and a gradual increase
in $\text{N}_2\text{O}$ from near-zero. CCSS-RS tracks the $\text{NH}_4$ decay
trajectory faithfully, maintaining well-calibrated intervals that tighten as the signal
stabilises, and correctly predicts the $\text{N}_2\text{O}$ increase from zero via the
hurdle mechanism. CCSS-SR tracks the decay trend with slightly wider intervals but
loses accuracy on the $\text{N}_2\text{O}$ timing. CCSS-G substantially underestimates
the $\text{N}_2\text{O}$ rise, because its Gaussian likelihood penalises deviations from zero
symmetrically, so the model anchors near zero rather than modelling the gradual increase.
Both Neural CDEs produce flat trajectories, unable to represent the multi-hour transient
that dominates this window (Figure~\ref{fig:window2}). Together, the two windows
illustrate that regime switching, heavy-tailed likelihoods, and the gain-weighted forcing
mechanism are not incremental refinements but qualitatively change the model's ability
to represent realistic process dynamics.

\section{Application-Oriented Case Studies for Operator Support}
\label{sec:case_studies}

\begin{figure*}[t]
\centering
\begin{tikzpicture}[
    font=\small,
    >=Latex,
    node distance=4mm and 4.5mm,
    stage/.style={
        draw=black!55,
        line width=0.9pt,
        rounded corners=5pt,
        align=center,
        fill=white,
        minimum height=2.0cm,
        text width=0.149\textwidth,
        inner sep=6pt
    },
    badge/.style={
        circle,
        draw=white,
        line width=0.9pt,
        text=white,
        font=\bfseries\footnotesize,
        minimum size=5.5mm,
        inner sep=0pt
    },
    arrow/.style={->, very thick, draw=black!60},
    title/.style={font=\bfseries\small},
    detail/.style={font=\footnotesize, text=black!75}
]
\node[stage, fill=cyan!8, draw=cyan!50!black] (context)
{\textbf{Recent Plant Context}\\[3pt]
{\footnotesize\color{black!75} Recent sensor history\\and plant state}};
\node[stage, fill=teal!8, draw=teal!55!black, right=of context] (plans)
{\textbf{Candidate Future Control Plans}\\[3pt]
{\footnotesize\color{black!75} Alternative setpoints\\or valve plans}};
\node[stage, fill=orange!12, draw=orange!70!black, right=of plans] (rollout)
{\textbf{Learned Simulator / CCSS-RS Rollout}\\[3pt]
{\footnotesize\color{black!75} Same exogenous forecast,\\different candidate plans}};
\node[stage, fill=gray!10, draw=black!55, right=of rollout] (compare)
{\textbf{Trajectory and Heuristic Comparison}\\[3pt]
{\footnotesize\color{black!75} Variables, bands,\\and screening scores}};
\node[stage, fill=green!10, draw=green!50!black, right=of compare] (review)
{\textbf{Operator or Engineer Review}\\[3pt]
{\footnotesize\color{black!75} Shortlist for deeper\\engineering analysis}};

\node[badge, fill=cyan!60!black, anchor=south west] at ([xshift=-1.5mm,yshift=-1.2mm]context.north west) {1};
\node[badge, fill=teal!65!black, anchor=south west] at ([xshift=-1.5mm,yshift=-1.2mm]plans.north west) {2};
\node[badge, fill=orange!85!black, anchor=south west] at ([xshift=-1.5mm,yshift=-1.2mm]rollout.north west) {3};
\node[badge, fill=black!65, anchor=south west] at ([xshift=-1.5mm,yshift=-1.2mm]compare.north west) {4};
\node[badge, fill=green!55!black, anchor=south west] at ([xshift=-1.5mm,yshift=-1.2mm]review.north west) {5};

\draw[arrow] (context) -- (plans);
\draw[arrow] (plans) -- (rollout);
\draw[arrow] (rollout) -- (compare);
\draw[arrow] (compare) -- (review);
\end{tikzpicture}
\caption{Conceptual workflow for using CCSS-RS as a learned simulator for scenario
screening and decision support. Recent plant context is paired with candidate future
control plans, rolled out under the learned simulator, compared through trajectories and
simple heuristics, and then reviewed by an operator or engineer before deeper analysis.}
\label{fig:workflow}
\end{figure*}
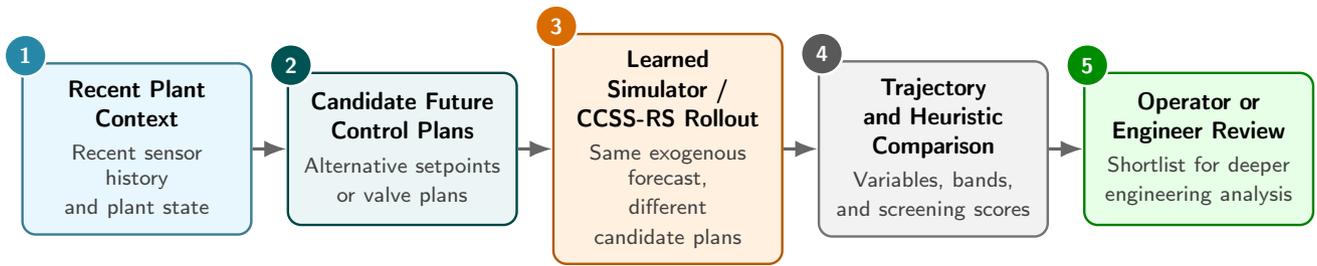

Benchmark accuracy alone does not show how a learned simulator would be used by a process
engineer. We therefore present four application-oriented case studies using the frozen
CCSS-RS checkpoint and \emph{test data only}. No retraining is performed, and the
benchmark results remain unchanged. Figure~\ref{fig:workflow} summarises the intended
scenario-screening workflow.

\subsection{Representative Window Selection Protocol}

To avoid arbitrary example selection, we scored all 11,839 fully observed test windows
using summary measures computed from the observed rollout only: $\text{NH}_4$ transient
magnitude and total variation, $\text{O}_2$ cycling intensity, $\text{N}_2\text{O}$ event
strength, and the number of phase changes. For the two what-if figures, we then
refined the ammonium-transient and oxygen-cycling candidates by restricting attention to
the top 1,000 windows for each behaviour type and selecting those with the best
observation-matched baseline rollout under the frozen CCSS-RS checkpoint, measured by a
weighted relative RMSE over the plotted states. The remaining $\text{N}_2\text{O}$-event
and phase-rich windows were selected using the original dynamics-only rule, with a minimum
start-index separation of 2,000 steps across the final four examples. The selected windows
therefore represent strong $\text{NH}_4$ transients, strong $\text{O}_2$ cycling,
pronounced $\text{N}_2\text{O}$ behaviour, and phase-rich operation, respectively.

This procedure ties the examples to distinct plant behaviours that an engineer might
realistically want to screen, while avoiding what-if figures whose baseline rollout is
already dominated by an obvious prediction failure.

\subsection{What-If Control Analysis Under Alternative Future Plans}

The first application question is whether the simulator can compare alternative future
control plans from the same recent operating context. For two representative windows
(an ammonium-transient window and an oxygen-cycling window), we keep the observed exogenous and categorical
future inputs fixed and replace only the continuous control plan at inference time. We
compare the observed plan against three moderate perturbations: $\text{O}_2$ setpoint
$+0.2$, $\text{O}_2$ setpoint $-0.2$, and valve opening $+10$ percentage points. The
purpose is not to claim causal plant behaviour, but to show how the learned simulator can
surface materially different rollout trajectories under alternative candidate plans.

\begin{figure*}[t]
\centering
\includegraphics[width=1.0\linewidth]{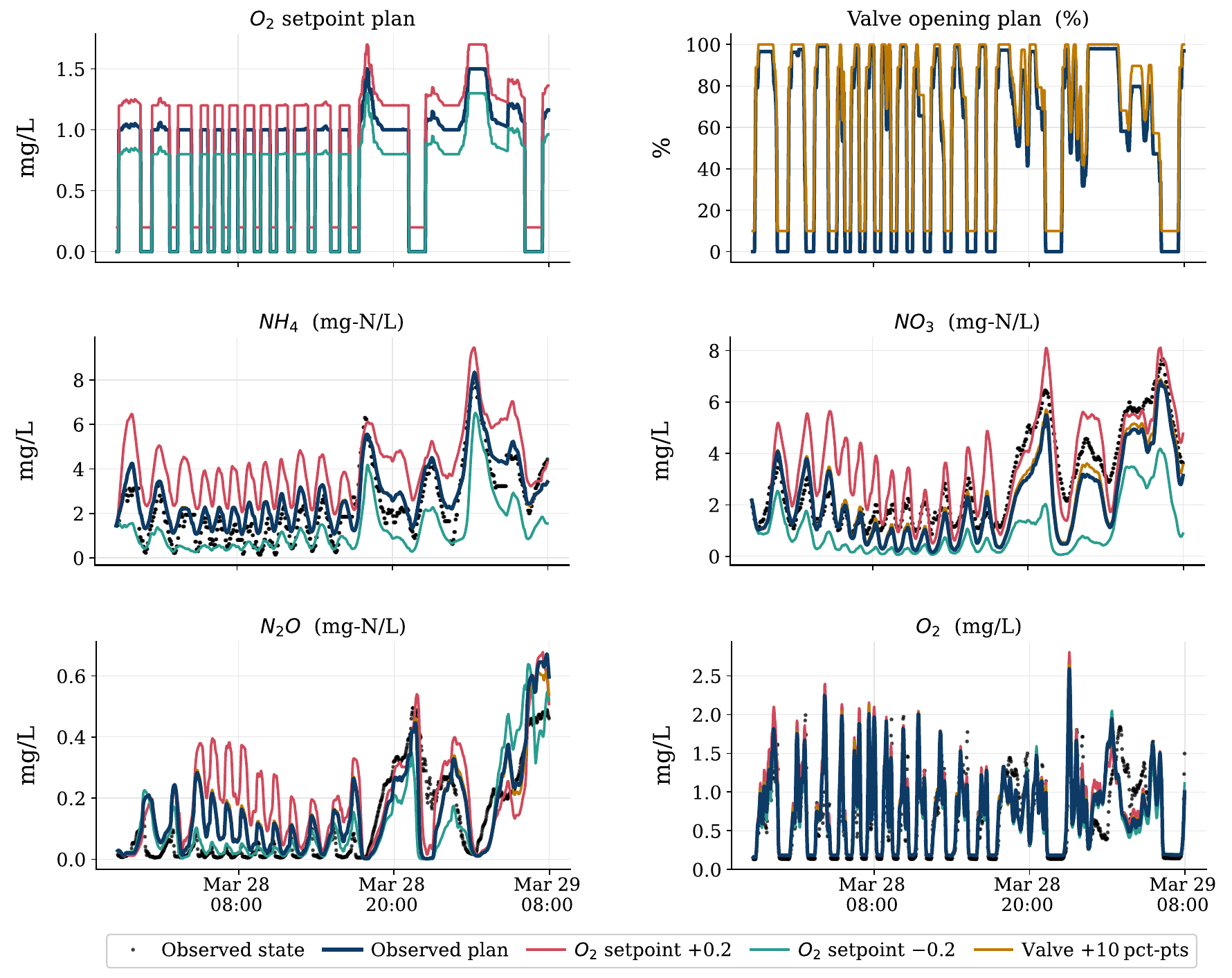}
\caption{What-if control comparison for the selected ammonium-transient test window. The
same historical context is rolled out under the observed future plan and three moderate
control perturbations. The learned simulator predicts substantially larger trajectory
separation under $\text{O}_2$ setpoint perturbations than under the tested valve
perturbation.}
\label{fig:case_whatif_nh4}
\end{figure*}

\begin{figure*}[t]
\centering
\includegraphics[width=1.0\linewidth]{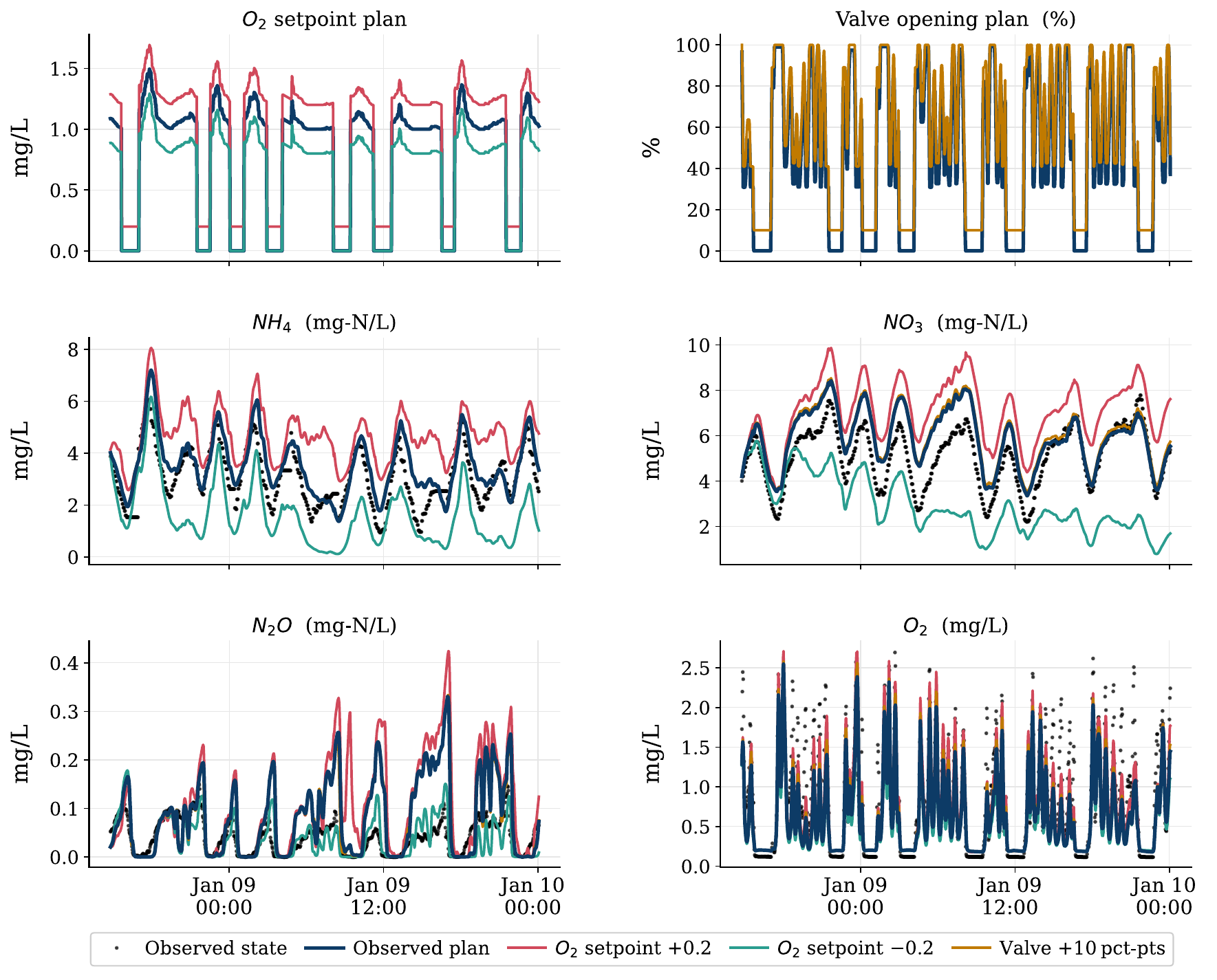}
\caption{What-if control comparison for the selected oxygen-cycling test window. The same
context produces clearly different $\text{NH}_4$ and $\text{N}_2\text{O}$ trajectories
under alternative future setpoint plans, illustrating how a learned simulator can support
relative plan comparison even when the directional response is regime dependent.}
\label{fig:case_whatif_o2}
\end{figure*}

Table~\ref{tab:what_if_summary} quantifies the trajectory differences relative to the
observed plan. The $\pm0.2$ setpoint perturbations produce much larger changes than the
$+10$ valve perturbation. In the ammonium-transient window, the setpoint-down scenario
shifts predicted $\text{NH}_4$ by $-1.070$ at $H=300$ and $-1.865$ at $H=1000$, whereas
the setpoint-up scenario shifts it by $+1.382$ and $+0.882$ respectively. In the
oxygen-cycling window, the corresponding deltas are $-1.731/+0.534$ at $H=300$ and
$-2.326/+1.410$ at $H=1000$. The tested valve perturbation produces only
$+0.033$ to $+0.065$ $\text{NH}_4$ change at $H=300$. These results do not validate the
perturbed plans as operationally correct, but they do show that the learned simulator is
sensitive enough to discriminate among plausible candidate control trajectories.

The non-monotone sign of the learned $\text{O}_2$-setpoint response should therefore be
read with process-regime caution rather than as a general aeration rule. In activated-sludge
systems, intermittent aeration, aerobic/anoxic switching, and dissolved-oxygen gradients can
change the local balance between nitrification, denitrification, and $\text{N}_2\text{O}$
formation pathways \citep{dotro2011intermittent,satoh2003oxygen,wunderlin2012mechanisms}.
We therefore
interpret these rollouts as regime-conditioned learned-simulator responses under the
observed exogenous plan, not as validated causal effects of raising or lowering aeration.

For these windows, $\text{O}_2$-setpoint changes are materially more consequential than
the tested valve perturbation for simulated $\text{NH}_4$ response.

\begin{table*}[t]
\centering
\caption{Scenario deltas relative to the observed plan for two representative windows.
Values are predicted state differences under the perturbed future plan.}
\label{tab:what_if_summary}
\small
\setlength{\tabcolsep}{5pt}
\begin{tabular}{llrrrr}
\toprule
\textbf{Window} & \textbf{Scenario} & \textbf{$\Delta \text{NH}_4$ @300} & \textbf{$\Delta \text{NH}_4$ @1000} & \textbf{$\Delta \text{N}_2\text{O}$ @300} & \textbf{$\Delta \text{N}_2\text{O}$ @1000} \\
\midrule
Ammonium-transient & $\text{O}_2$ setpoint $+0.2$ & $+1.382$ & $+0.882$ & $+0.207$ & $-0.089$ \\
Ammonium-transient & $\text{O}_2$ setpoint $-0.2$ & $-1.070$ & $-1.865$ & $-0.060$ & $-0.075$ \\
Ammonium-transient & Valve $+10$ pct-pts        & $+0.065$ & $+0.028$ & $+0.009$ & $-0.059$ \\
Oxygen-cycling     & $\text{O}_2$ setpoint $+0.2$ & $+0.534$ & $+1.410$ & $+0.005$ & $+0.051$ \\
Oxygen-cycling     & $\text{O}_2$ setpoint $-0.2$ & $-1.731$ & $-2.326$ & $-0.006$ & $-0.064$ \\
Oxygen-cycling     & Valve $+10$ pct-pts        & $+0.033$ & $-0.020$ & $+0.001$ & $+0.002$ \\
\bottomrule
\end{tabular}
\end{table*}

\subsection{Simulator-Assisted Screening of Candidate Control Trajectories}

The second application question is whether the simulator can rank multiple candidate
future plans before deeper engineering review. We construct eight constrained variants
of the observed future control sequence for the selected phase-transition-rich window,
each targeting a distinct operational intention: \textbf{setpoint $\pm$0.1} shifts the
$\text{O}_2$ setpoint uniformly by $\pm$0.1\,mg/L throughout the horizon, testing
whether a marginally lower or higher aeration target improves effluent quality;
\textbf{smoothed setpoint} replaces abrupt shift-transition step-changes with a
moving-average profile (21-step kernel, $\approx$20\,min), reducing biomass stress;
\textbf{front-load $\pm$0.2} applies a $\pm$0.2\,mg/L setpoint push only in the
first third of the horizon ($\approx$4\,h) then reverts to the observed schedule,
testing whether an early aeration adjustment has lasting downstream effects; and
\textbf{valve $\pm$5} shifts the valve opening, an independent actuator controlling
aeration flow rate, by $\pm$5\,percentage points throughout, independently of the
setpoint target. Each candidate is evaluated with four transparent surrogate criteria:
mean predicted $\text{NH}_4$ over the horizon (weight 0.40), the 95th percentile of
predicted $\text{N}_2\text{O}$ (0.25), a soft $\text{O}_2$-band violation penalty
(0.20), and a control-deviation penalty relative to the observed plan (0.15). Scores
are min-max normalised and combined; lower is better. This is framed strictly as
\emph{scenario screening support}, not control optimisation.

\begin{figure*}[t]
\centering
\includegraphics[width=1.0\linewidth]{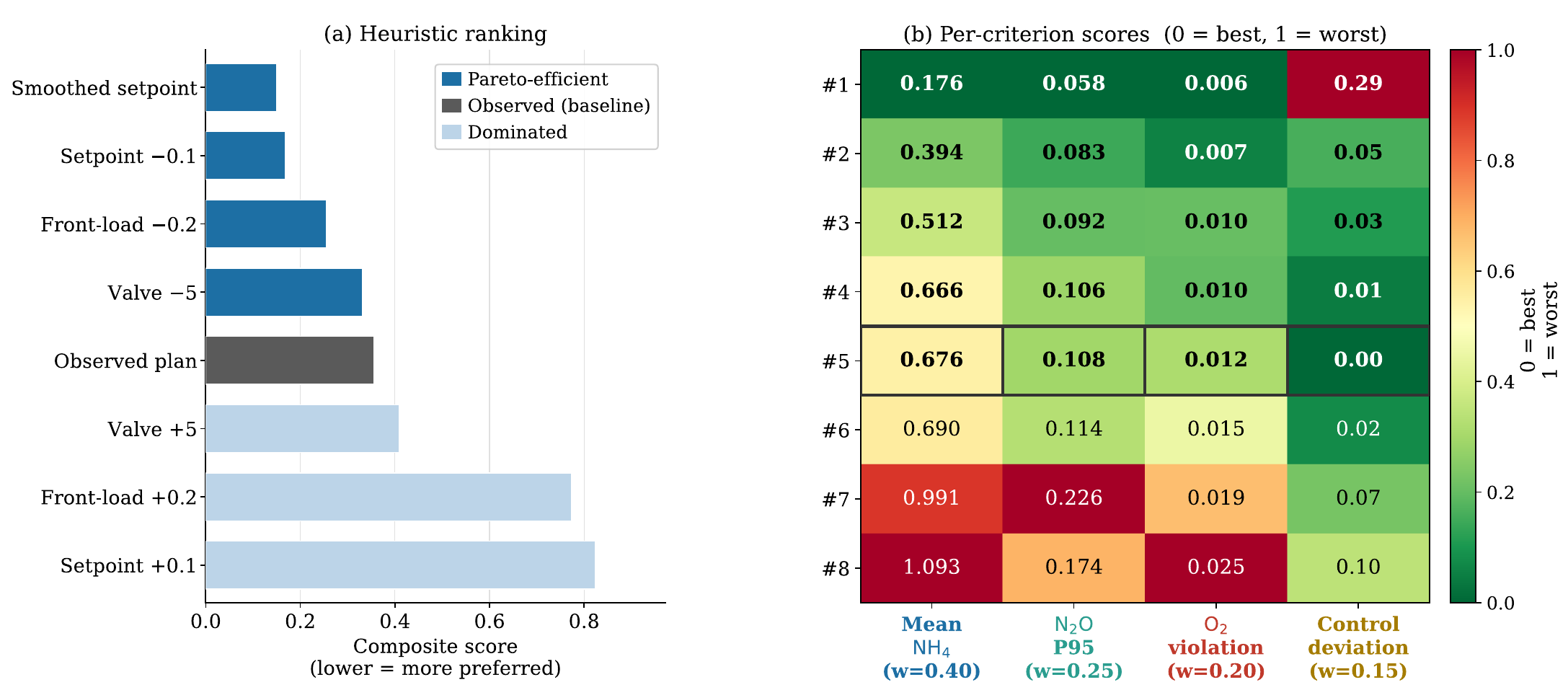}
\caption{Candidate-plan screening results for the selected phase-rich test window.
(a)~Composite heuristic ranking (lower score = more preferred); Pareto-efficient plans
are shown in blue, the observed plan in grey, and dominated plans in light blue.
(b)~Per-criterion normalised scores (0 = best, 1 = worst) with raw values annotated.
Bold text marks Pareto-efficient plans; the observed plan row is highlighted with a
border. The smoothed setpoint plan dominates on NH$_4$ and N$_2$O criteria while
trading off control deviation.}
\label{fig:screening}
\end{figure*}

Figure~\ref{fig:screening} and Table~\ref{tab:screening} present the results. The
smoothed setpoint plan ranks first, achieving the lowest mean $\text{NH}_4$ (0.176)
and $\text{N}_2\text{O}$ P95 (0.058) by a wide margin, at the cost of the highest
control deviation (0.290), since the smoother profile departs most from the actual
observed schedule. The observed plan ranks fifth of eight. All five lowest-scoring
plans are Pareto-efficient; the two positive-setpoint perturbations are dominated on
every relevant criterion. The learned simulator thus narrows the
candidate set and exposes trade-offs before more expensive controller testing or
mechanistic analysis.

\begin{table}[t]
\centering
\caption{Heuristic screening ranking of eight candidate control plans for the selected
phase-rich test window. Criteria and composite weights: mean predicted
$\text{NH}_4$ (0.40), 95th-percentile predicted $\text{N}_2\text{O}$ (0.25),
soft $\text{O}_2$-band violation (0.20), and control deviation from the observed plan
(0.15). Composite scores are the weighted sum of min-max-normalised criteria; lower is
better. Plans marked $\dagger$ are Pareto-efficient.}
\label{tab:screening}
\small
\setlength{\tabcolsep}{3.5pt}
\resizebox{\columnwidth}{!}{%
\begin{tabular}{clccccc}
\toprule
\textbf{Rank} & \textbf{Scenario} & \textbf{Mean} & \textbf{N$_2$O} & \textbf{O$_2$} & \textbf{Ctrl} & \textbf{Score}\\
 & & \textbf{NH$_4$} & \textbf{P95} & \textbf{Viol.} & \textbf{Dev.} & \\
\midrule
1 & Smoothed setpoint$^\dagger$    & 0.176 & 0.058 & 0.006 & 0.290 & 0.150 \\
2 & Setpoint $-0.1$$^\dagger$       & 0.394 & 0.083 & 0.007 & 0.048 & 0.168 \\
3 & Front-load $-0.2$$^\dagger$     & 0.512 & 0.092 & 0.010 & 0.033 & 0.255 \\
4 & Valve $-5$$^\dagger$            & 0.666 & 0.106 & 0.010 & 0.014 & 0.332 \\
5 & \textit{Observed}$^\dagger$     & 0.676 & 0.108 & 0.012 & 0.000 & 0.355 \\
6 & Valve $+5$                      & 0.690 & 0.114 & 0.015 & 0.022 & 0.408 \\
7 & Front-load $+0.2$               & 0.991 & 0.226 & 0.019 & 0.067 & 0.773 \\
8 & Setpoint $+0.1$                 & 1.093 & 0.174 & 0.025 & 0.100 & 0.823 \\
\bottomrule
\end{tabular}}
\end{table}

In this screening setup the learned simulator functions as a transparent filter for
ranking candidate plans before deeper controller or mechanistic review. The criteria
weights and thresholds are configurable and can be adapted by the process engineer to
reflect local regulatory priorities (e.g.\ up-weighting the $\text{N}_2\text{O}$ term
if greenhouse-gas mitigation is the primary concern).

\begin{figure*}[t]
\centering
\includegraphics[width=1.0\linewidth]{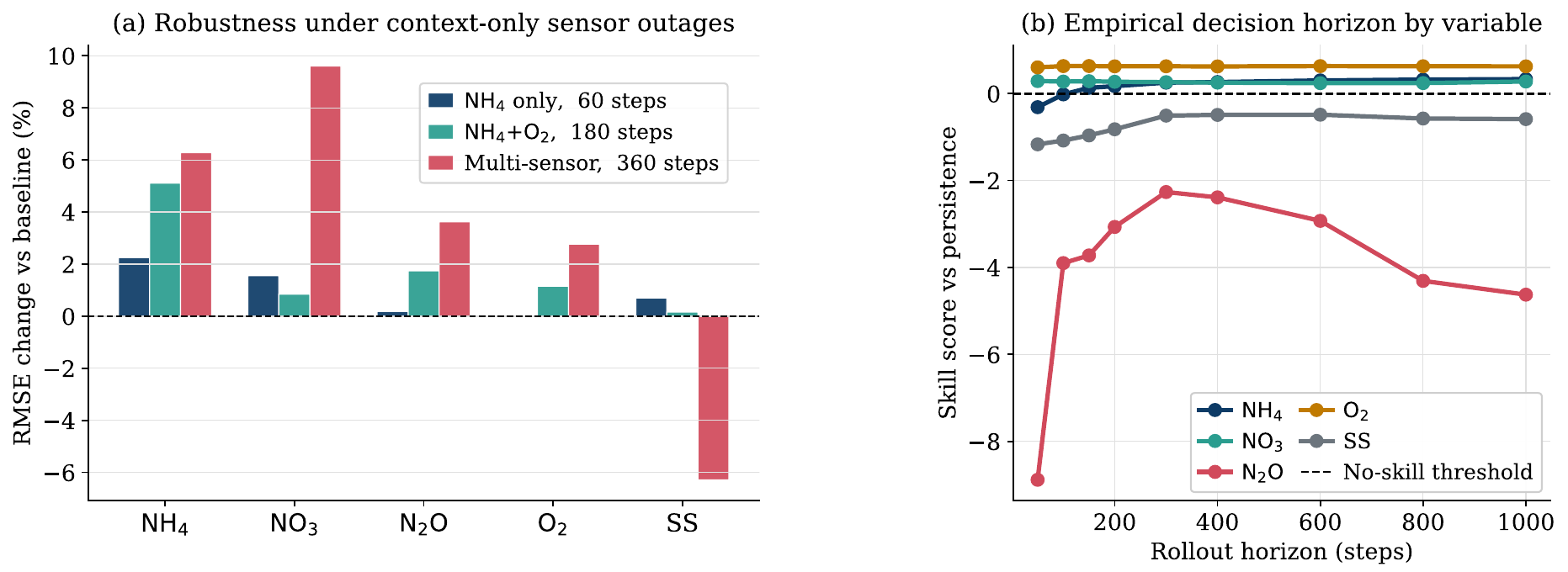}
\caption{Operator-support summary from the robustness and decision-horizon analyses.
(a)~Per-variable RMSE change under context-only sensor outages (three conditions,
64 test windows); boxplots below show the distribution of overall window-level RMSE
ratios.
(b)~Skill against a persistence baseline as a function of rollout horizon (same 64
windows). $\text{NH}_4$, $\text{NO}_3$, and $\text{O}_2$ retain positive skill across
the full 1000-step rollout; $\text{N}_2\text{O}$ and SS do not.}
\label{fig:operator_support}
\end{figure*}

\subsection{Robustness Under Recent Sensor Outages}

A practical learned simulator must remain useful when recent context measurements are
incomplete. We therefore mask additional \emph{context-only} observations over 64 random
fully observed test windows and compare the resulting rollout error against the same
windows without extra masking. Three outage conditions are tested: the final 60 context
steps of $\text{NH}_4$, the final 180 steps of $\text{NH}_4$ and $\text{O}_2$, and a
longer 360-step outage affecting $\text{NH}_4$, $\text{NO}_3$, $\text{N}_2\text{O}$, and
$\text{O}_2$ together.

The short single-sensor outage increases $\text{NH}_4$ RMSE by 2.2\% and
$\text{NO}_3$ RMSE by 1.6\%, with near-zero change for $\text{N}_2\text{O}$ and
$\text{O}_2$. The longer dual-sensor outage raises $\text{NH}_4$ RMSE by 5.1\% and
$\text{N}_2\text{O}$ RMSE by 1.7\%. Under the longest multi-sensor outage, the error
increase remains bounded at 6.3\% for $\text{NH}_4$, 9.6\% for $\text{NO}_3$, 3.6\% for
$\text{N}_2\text{O}$, and 2.8\% for $\text{O}_2$. This does not eliminate the value of
recent sensing, but it suggests that the continuous-time, missingness-aware setup still
supports useful simulation when recent plant history is partially unavailable
(Figure~\ref{fig:operator_support}(a)).

Operationally, these outage results suggest that short recent sensor losses need not
invalidate scenario screening, whereas longer multi-sensor outages should reduce
confidence in the rollout.

\subsection{From Benchmark Horizon to Practical Decision Horizon}

Finally, we translate the $H=1000$ benchmark into a more operator-oriented notion of
decision horizon. Over the same 64-window sample, we compare cumulative CCSS-RS RMSE
against a persistence baseline that holds the final context belief state constant. We then
define an empirical ``useful horizon'' as the longest horizon at which the model retains
positive skill over persistence while staying within one training-scale unit of normalised
RMSE. This criterion is intentionally conservative and site specific.

Under this criterion, $\text{NH}_4$, $\text{NO}_3$, and $\text{O}_2$ retain positive
skill throughout the full 1000-step rollout, corresponding to approximately 33.3 hours at
the median test-set sampling interval. $\text{N}_2\text{O}$ and SS do not satisfy the
criterion on this sample, indicating that benchmark competence at extreme horizon should
not be interpreted as uniform operational usefulness across all variables. This kind of
variable-specific horizon interpretation is more actionable for operator support than a
single aggregate benchmark score (Figure~\ref{fig:operator_support}(b)).

In practice, the simulator is more informative for $\text{NH}_4$, $\text{NO}_3$,
and $\text{O}_2$ scenario screening across long horizons than for $\text{N}_2\text{O}$
or SS.

Taken together, the four case studies demonstrate that a frozen CCSS-RS checkpoint
provides concrete operator-support capabilities: it surfaces materially different
trajectory outcomes under alternative future control plans, narrows a set of candidate
plans to a short-list with transparent configurable criteria, remains useful under
realistic short-term sensor outages, and identifies which variables carry decision-relevant
information at which planning horizons. Each of these capabilities is available with no
model retraining and no process model expertise from the end user. The combined picture
supports a deployment scenario in which the simulator functions as a fast digital-twin
screening layer upstream of any field actuation, reviewed by an operator before decisions
are made.

\section{Ablation Studies}
\label{sec:ablation}

\begin{figure}
    \centering
    \includegraphics[width=1.0\linewidth]{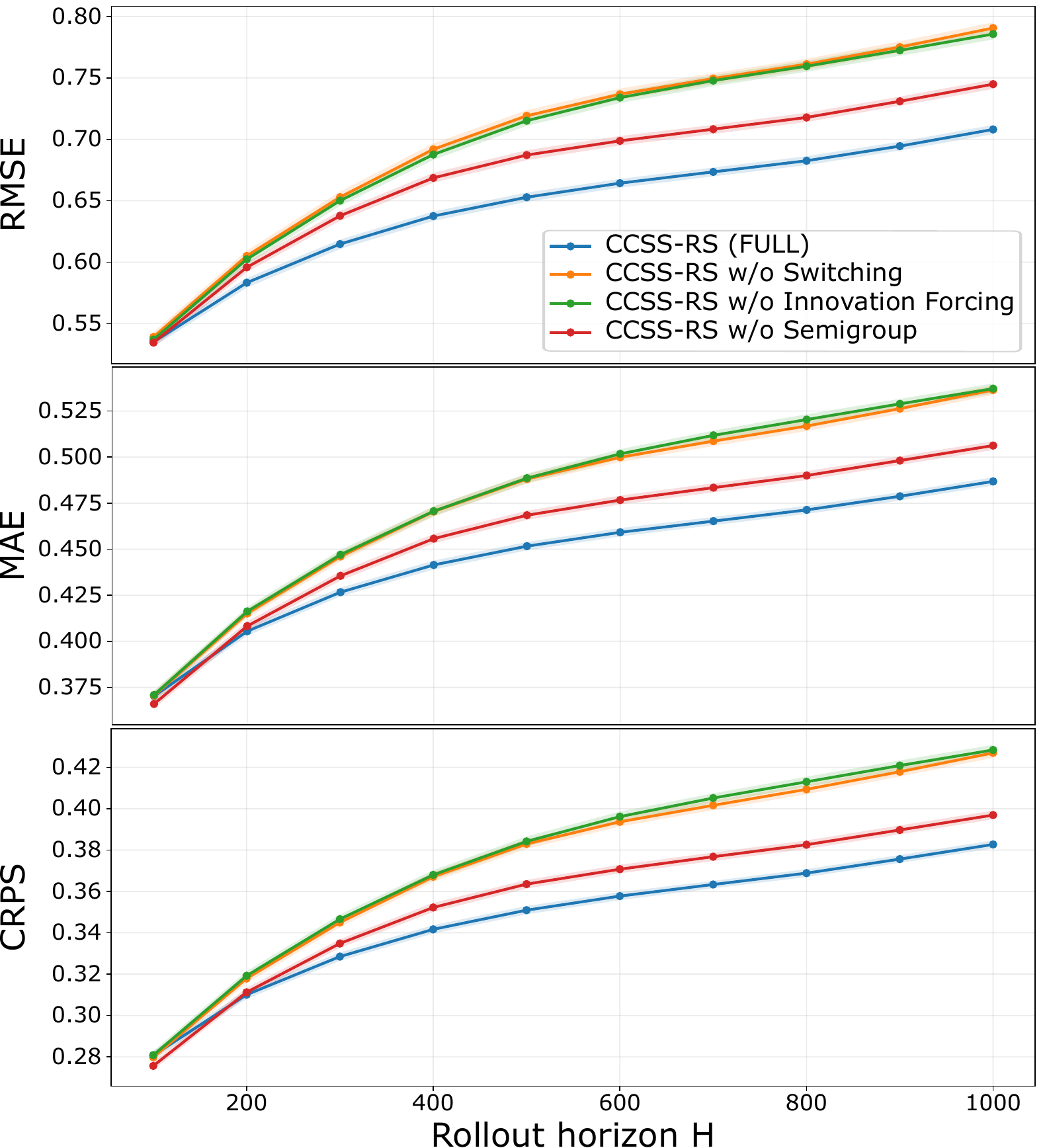}
    \caption{RMSE (top), MAE (middle), and CRPS (bottom) vs.\ horizon $H$ for CCSS-RS
    and three ablated variants across three training seeds. All three components contribute
    increasingly with horizon length, but with distinct growth profiles.}
    \label{fig:ablation}
\end{figure}

We construct three ablated variants from CCSS-RS (FULL): removing innovation forcing
(the gain-weighted driver forcing mechanism is disabled, leaving the base scan dynamics
without step-level control injection); removing semigroup consistency regularisation;
and removing regime switching (reducing to $K=1$). Each variant is trained three times
with seeds (41, 42, 43) to ensure results are not due to random initialisation.
Table~\ref{tab:ablation} and Figure~\ref{fig:ablation} present results over 5,000 windows.

\begin{table}[h]
\centering
\caption{Ablation results at $H=1000$ (mean over 5,000 windows $\times$ 3 seeds; lower
is better).}
\label{tab:ablation}
\small
\resizebox{\columnwidth}{!}{%
\begin{tabular}{lcccc}
\toprule
\textbf{Model} & \textbf{RMSE} & \textbf{$\Delta$RMSE} & \textbf{CRPS} & \textbf{$\Delta$CRPS} \\
\midrule
CCSS-RS (FULL)                 & $\mathbf{0.708}$ &  -         & $\mathbf{0.383}$ &  -         \\
CCSS-RS w/o Switching          & $0.745$          & $+5.2\%$  & $0.397$          & $+3.6\%$  \\
CCSS-RS w/o Innovation Forcing & $0.786$          & $+11.0\%$ & $0.428$          & $+11.8\%$ \\
CCSS-RS w/o Semigroup          & $0.791$          & $+11.7\%$ & $0.427$          & $+11.5\%$ \\
\bottomrule
\end{tabular}}
\end{table}

Three findings emerge. First, innovation forcing and semigroup consistency make almost
identical contributions at $H=1000$ ($+11.0\%$ and $+11.7\%$ RMSE degradation), yet
they address distinct failure modes: innovation forcing ensures that gain-weighted control
and exogenous driver signals are actively injected into the latent trajectory at each
step, preventing the base scan dynamics from under-responding to prescribed future inputs;
semigroup consistency prevents accumulation of step-size-dependent discretisation errors.
Critically, both effects are negligible at $H=100$ and grow substantially with rollout
length, confirming that they target failure modes specific to long-horizon simulation
rather than one-step prediction, a non-trivial finding that justifies their inclusion in
MPC-focused architectures.

Second, regime switching has the smallest isolated effect at $H=1000$ ($+5.2\%$ RMSE)
but a qualitatively different horizon profile: the gap is essentially zero at $H=100$,
remaining negligible through $H\approx300$, then widening monotonically beyond $H=400$.
This delayed onset is mechanistically coherent: mode errors during one ASP phase
compound as the process transitions to subsequent phases, so the cost of misidentification
accumulates across multiple cycle lengths rather than manifesting immediately.

Third, comparing the ablation ($+5.2\%$ from removing switching) against the main result
($+44\%$ for CCSS-SR vs.\ CCSS-RS) reveals that the larger gap in the main comparison
reflects cumulative differences across multiple design dimensions: encoder architecture
(TCN vs.\ IIR), parameter allocation, and training dynamics. The two experiments are
complementary: ablations show the marginal cost of each component in isolation;
the main comparison shows the practical consequence of the full design space.

\section{Discussion}
\label{sec:discussion}

\subsection{Practical Placement in Industrial Workflows}

The case studies clarify a realistic operator workflow for the learned simulator. First,
an engineer selects a recent context window and proposes a small set of future control
plans. Second, the simulator rolls out each plan under the same exogenous forecast and
returns predicted trajectories for variables of concern such as $\text{NH}_4$,
$\text{NO}_3$, $\text{N}_2\text{O}$, and $\text{O}_2$. Third, the plans are compared by
transparent surrogate criteria or visualised side by side before any field change is
considered. In this role, the simulator functions as a decision-support filter: it helps
identify which candidate plans merit closer attention, which windows are highly sensitive
to setpoint changes, and which variables appear robust or fragile at a given horizon.

In practice, this tool is best positioned upstream of any field actuation: for example,
as an offline screening aid before a shift or as a fast comparator when tuning MPC
setpoints and reviewing a small set of candidate plans. It is not intended to replace
calibrated mechanistic models; rather, it can be used alongside them to narrow candidate
actions or flag materially different predicted futures before deeper engineering review.
In that sense, CCSS-RS is better viewed as a learned simulator usable as a digital-twin
component, not as a deployed digital twin.

The outage analysis is equally important for practical realism. Short recent sensor
outages raised RMSE only modestly, suggesting that the continuous-time, mask-aware design
retains useful context even when the latest measurements are incomplete. The decision-horizon analysis further shows that utility is variable specific: $\text{NH}_4$,
$\text{NO}_3$, and $\text{O}_2$ retain positive skill against persistence across the full
rollout, whereas $\text{N}_2\text{O}$ and SS do not under the adopted criterion. For
engineering use, this means the simulator is better suited to some operational questions
than others, and those boundaries should be reported explicitly.

\subsection{Deployment Considerations}

A practical deployment pathway for CCSS-RS consists of four steps. \textbf{Data
integration:} the model consumes raw historian data from plant SCADA/DCS systems, with no
external resampling or imputation required beyond the native data pipeline. Inference from
a 512-step context window (approximately 8--10 hours of data at typical WWTP sampling
rates) takes approximately 1.2 seconds on a single CPU for the full 1000-step rollout,
making pre-shift or mid-shift screening feasible. The 26.6\,MB checkpoint is deployable on
standard industrial servers without specialised hardware. \textbf{Model training:} a full
training run (25 epochs on two years of data) requires 6.5 minutes on a single NVIDIA RTX
PRO 6000 Blackwell GPU; periodic retraining after seasonal change, maintenance, or major
configuration updates is therefore computationally practical. \textbf{Interface:} output
trajectories and predictive intervals can be served through a thin REST API to existing
operator dashboards or as tabular exports compatible with standard process-historian
tools. \textbf{Governance:} because the simulator is framed as a screening aid rather than
a controller, operators retain authority over actuation decisions and can use the model as
an advisory layer before any closed-loop deployment is considered. These characteristics
support integration as an offline advisory component in WWTP operational support systems,
while stopping short of automated control.

\subsection{Design Principles for Engineering-Oriented Learned Simulators}

Four design principles emerge from the benchmark and case-study results, each with direct
implications for practitioners building similar tools in other industrial domains.

First, \emph{future controls must be modelled as privileged drivers}, not as
undifferentiated covariates. The engineering task is conditional simulation under a
specified future plan, and blending control signals with process observations in a single
input stream loses the control-state distinction that makes what-if queries possible.

Second, \emph{irregular sampling and missingness must be first-class citizens}.
Forward-filling 43\% missingness over 1000 steps introduces systematic false confidence
and breaks like-for-like evaluation. Architectures that natively consume sparse irregular
observations avoid this avoidable source of bias and are better matched to realistic plant
conditions.

Third, \emph{long-horizon stability is a distinct failure mode from one-step fit}. The
semigroup and forcing ablations show that components designed for rollout stability
contribute negligibly at $H=100$ but critically at $H=1000$. Optimising validation loss
at short horizons will not discover or validate these components.

Fourth, \emph{application value comes from comparison, not only prediction}. The
what-if and screening studies are useful precisely because the simulator supports relative
assessment of candidate plans under a common historical context. Absolute trajectory
accuracy matters less than the simulator's ability to rank plans consistently; this is a
different evaluation target than conventional benchmark scores and should be tested
explicitly during system validation.

\subsection{Limitations and Future Directions}

Evaluation is confined to a single facility; cross-facility generalisation remains
unproven, and transfer to plants with different biological configurations or sensor suites
would require validation and likely fine-tuning. We do not compare against a calibrated
ASM-family mechanistic model, and do not claim the learned simulator should replace one
where mechanism-level process understanding is required. The what-if perturbation studies
must be interpreted with care: because the model is trained on observational plant data
rather than interventional experiments, scenario responses are learned associations, not
causal guarantees. This is precisely why the application framing is scenario
\emph{screening support} rather than control synthesis, and why operator review remains an
explicit step in the workflow. Concept drift from biofilm growth, planned maintenance, or
seasonal process change is not explicitly modelled; periodic retraining or lightweight
online adaptation would be needed for sustained production deployment. Finally, while
regime assignments improve interpretability, within-regime dynamics remain opaque, which
may limit regulatory acceptance in jurisdictions requiring explainable decision support.

The conceptual parallel with learned world models in reinforcement learning
\citep{ha2018world,hafner2023dreamerv3} is instructive: structured latent dynamics can
enable planning under known action sequences, although those approaches assume
dense regular observations and closed-loop episodic interaction rather than sparse
open-loop industrial operation. Hybrid architectures combining learned dynamics with
mechanistic process constraints, interpretable uncertainty quantification, online
parameter adaptation, and multi-site transfer are the natural next directions for
increasing the operational readiness and regulatory acceptance of tools like CCSS-RS.

\section{Conclusion}
\label{sec:conclusion}

We presented CCSS-RS as a data-driven open-loop simulator for digital-twin operator
decision support in wastewater treatment, designed to work under the data conditions of
real full-scale facilities without signal resampling, gap-filling, or process model
recalibration. The benchmark results confirm that explicit control-state separation,
gain-weighted forcing, regime-aware dynamics, semigroup consistency, and heavy-tailed
probabilistic outputs each contribute substantially and are together responsible for
reliable long-horizon rollout: CCSS-RS reaches RMSE 0.696 and CRPS 0.349 at $H=1000$ across 10,000 evaluation
windows, outperforming the most directly compatible external baselines as well as simplified
internal variants. Ablations show that the main gains are specific to long-horizon
stability rather than one-step prediction, confirming relevance for MPC-scale planning
horizons.

The four application case studies translate those benchmark gains into operational value.
The simulator distinguishes among alternative future control plans from the same context,
ranks candidate plans with transparent configurable criteria, maintains useful rollouts
under moderate sensor outages, and identifies which state variables retain practical
decision-horizon value. These demonstrations support a concrete deployment pathway:
CCSS-RS can be integrated as an offline advisory layer in WWTP operational systems,
requiring no process model expertise from the end user and keeping human review upstream
of any closed-loop control change.

The clearest limitation is evaluation on a single facility; cross-facility
generalisation, hybrid integration with mechanistic models, and online adaptation to
concept drift remain open research directions. Within those bounds, however, CCSS-RS
represents a practically viable learned simulator for digital-twin-style scenario
screening in industrial wastewater treatment: most valuable when its predictions are
used to compare candidate plans rather than to dictate them, and most informative for
the $\text{NH}_4$, $\text{NO}_3$, and $\text{O}_2$ variables that retain positive skill
across the full planning horizon.

\section*{CRediT authorship contribution statement}

Gary Simethy: Conceptualization, Methodology, Software, Writing -- original draft.
Daniel Ortiz Arroyo: Supervision, Methodology, Writing -- review \& editing.
Petar Durdevic: Supervision, Resources, Writing -- review \& editing.

\section*{Declaration of competing interest}

The authors declare that they have no known competing financial interests or personal
relationships that could have appeared to influence the work reported in this paper.

\section*{Funding}

This research was supported by Aalborg University and Helix Lab in Denmark under the
Novo Nordisk Fonden through project grant number 224611.

\section*{Acknowledgements}

The authors acknowledge support from Aalborg University and Helix Lab in Denmark.

\section*{Data and code availability}

The benchmark dataset used in this study is publicly available
\citep{hansen2024avedoredata}. Code used for model training, evaluation, and case-study
analysis is available from the authors upon reasonable request.

\section*{Declaration of generative AI and AI-assisted technologies in the manuscript preparation process}

During the preparation of this work, OpenAI's ChatGPT was used for language editing. The authors reviewed and edited the content as
needed and take full responsibility for the content of the published article.

\bibliographystyle{cas-model2-names}
\bibliography{cas-refs}

\end{document}